\pdfoutput=1
\documentclass[11pt]{article}
\usepackage{acl}
\usepackage{times}
\usepackage{latexsym}
\usepackage[T1]{fontenc}
\usepackage[utf8]{inputenc}
\usepackage{microtype}
\IfFileExists{inconsolata.sty}{\usepackage{inconsolata}}{}
\usepackage{graphicx,booktabs,array,amsmath,amssymb,caption,url,float}
\usepackage{xurl}
\usepackage{etoolbox}
\IfFileExists{dblfloatfix.sty}{}{}
\extrafloats{400}

\newsavebox{\tblbox}
\title{Attention Temperature Confounds Zero-Shot Context Extension}
\author{Ruben Navasardyan\thanks{\hspace{0.2em}Correspondence: \texttt{rnavasar@uwaterloo.ca}.} \\ Machine Learning and Optimization Laboratory, EPFL \\ Department of Applied Mathematics, University of Waterloo}
\begin{document}
\maketitle
\begin{abstract}
A training-time attention change is usually judged for zero-shot context extension against a dense baseline beyond the training length at the default temperature, $\tau$~=~1. The outcome depends on how miscalibrated the baseline is there: the share of perplexity it recovers when its logits are sharpened by one scalar. Under the standard recipe (fp32 master weights), an 8k-trained baseline is 8--15\% miscalibrated at 30k in three seed-paired 365M twins; in one, this makes the block-sparse-trained twin look 7\% better, which calibration removes. In pure-bf16 training without master weights the miscalibration is several times larger. An AdamW step ($\beta_{2}$ = 0.95) on a normalisation gain initialised at 1.0 is below half the bf16 spacing at peak learning rates under 1.7$\times$10$^{-3}$, so the gain never moves (22 checkpoints). The baseline is 37--58\% miscalibrated, and block-sparse pretraining, attention dropout and ALiBi look 6--35\% better. Calibrating one temperature per model and length removes all three advantages, leaves ALiBi 5\% behind RoPE and keeps part of the gain of a positive control, NTK-aware scaling. Unfreezing the gains reverses the block-sparse advantage at 365M and removes 71--89\% of it at 1.2B. We recommend reporting extension results at calibrated temperatures and checking that normalisation gains moved.

\end{abstract}
\section{Introduction}\label{introduction}

Zero-shot context extension asks how a model trained at one sequence length behaves at a longer one. A training-time change to attention is usually judged by comparing it with a dense baseline beyond the training length at the default attention temperature, $\tau$~=~1. That comparison has a hidden variable. Beyond the training length the softmax spreads attention over more keys than the logit scale was learned for, entropy rises, and sharpening the logits by one scalar recovers much of the loss \citep{chiang2022,su2021,peng2024,nakanishi2025}. We call the relative perplexity drop from $\tau$~=~1 to the best temperature the baseline's \emph{miscalibration} (a property of its logit scale, not probability calibration). A training-time change that merely tolerates $\tau$~=~1 better than the baseline then looks as if it extrapolates, by up to the baseline's miscalibration. This paper measures that confound under the standard training recipe (fp32 master weights, bf16 compute), shows that one precision setting makes it several times larger, and gives a check that removes it.

The vehicle is a set of seed-paired twins: a Dense model and a \emph{Mixed} model that trains under random block-sparse attention masks and is otherwise identical (§3; the name refers to the mixture of mask budgets, not to mixed precision). Under the standard recipe, in three 365M pairs, the dense baseline is 8--15\% miscalibrated at 30k, and in one pair that is enough for Mixed to look 7\% better at 30k, although the two arms agree to within 0.03 log-perplexity once both are calibrated and Mixed is 2--3\% worse within the training span (§4.4).

In pure-bf16 training without fp32 master weights, the AdamW update to a normalisation gain initialised at 1.0 is smaller than half the bf16 spacing at any peak learning rate below about 1.7$\times$10$^{-3}$ ($\beta_{2}$ = 0.95), so under round-to-nearest the gain never moves (§4.2). Every QK-normalisation gain in the 22 checkpoints of our eleven pure-bf16 Dense/Mixed pairs (365M, 1.2B and 3B) is exactly 1.0, so neither arm could learn an attention logit scale. The baseline's miscalibration at 30k is then 37\% at 365M and 50--58\% at 1.2B and 3B, and against it Mixed looks 6--31\% better, ALiBi 35\% better and, among the control arms we trained, attention dropout 8\% better and fixed sparse masks 30--36\% worse (§4.1). One calibrated temperature per model and length removes all of these advantages (ALiBi ends 5\% behind RoPE) and keeps part of the gain of a positive control, NTK-aware scaling. Unfreezing only the QK gains removes most of the amplified effect: Mixed falls behind at 365M, 71--89\% of its default advantage disappears at 1.2B, and ALiBi falls behind RoPE even at $\tau$~=~1 (§4.3). The standard recipe improves both arms within the span but leaves the baseline more miscalibrated beyond it than the learnable-gain recipe does (§4.4). At seeds 42 and 43, the two trained under both recipes, the frozen-gain baseline's miscalibration is three to five times the standard-recipe one in log-perplexity; why the frozen scale costs Dense more than Mixed is not identified.

Our contributions are three. (i) A measurement of how much the default-temperature comparison misstates a training-time attention change, across eleven twin pairs, four control arms and ALiBi, with the three precision recipes placed on one scale (Table 3). (ii) A check, one calibrated temperature per model and length with a positive control and a stated resolution, that detects the confound under either recipe without knowing its cause. (iii) The bf16/AdamW freeze condition, its learning-rate threshold, and its reversal by retraining at two scales.

\section{Related work}\label{related-work}

\textbf{Context extension and attention temperature.} Positional interpolation \citep{chen2023}, NTK-aware scaling \citep{bloc972023,peng2024}, LongRoPE \citep{ding2024} and continued pretraining \citep{xiong2024,fu2024} extend RoPE-based models \citep{su2024} beyond their window; training-free methods remap positions or restrict attention at inference \citep{han2024,jin2024,an2024}. Several works identify softmax entropy growth with key count as a failure mode and scale the logits by a length-dependent factor \citep{chiang2022,su2021,peng2024,nakanishi2025} or adapt the temperature to the attention entropy \citep{velikovi2025}; \citet{wang2024} tie the length-generalisation failure of a training-time change, NoPE \citep{kazemnejad2023}, to distracted attention and extend it by tuning per-head temperatures. None, to our knowledge, asks whether a training-time attention change can produce an apparent extension gain against a baseline that is merely miscalibrated, or how the training precision policy sets the size of that effect. Position-resolved evaluation is in the spirit of \citet{liu2024}; the seed-paired design is motivated by \citet{dodge2020} and the significance tests follow \citet{dror2018}.

\textbf{Sparse attention.} Fixed sparse patterns \citep{child2019,beltagy2020,zaheer2020}, content-based routing \citep{kitaev2020,roy2021} and natively trainable sparse architectures \citep{yuan2025,lu2025,gao2024,zhao2025} train and deploy with the same operator; inference-time sparse attention restricts or compresses what a dense-trained model attends to \citep{xiao2024,zhang2023,tang2024,nawrot2026}. Our Mixed arm perturbs training with random block masks and budgets, like stochastic structural regularisers \citep{fan2020,huang2016} and randomised positional encodings \citep{ruoss2023}, and is deployed here as an ordinary dense transformer. We use it only to measure the confound and do not propose it as a method.

\textbf{Low-precision training.} Training with an fp32 master copy of the weights is standard because updates below a parameter's half-precision resolution are otherwise lost (shown for fp16 by \citealp{micikevicius2018}); pure-bf16 training stalls parameters whose updates are small relative to their magnitude, which stochastic rounding or Kahan summation repair \citep{zamirai2020}; gains initialised at 1.0 are the most exposed. QK-normalisation \citep{henry2020} bounds the attention logits and is used at scale with a learnable gain to prevent logit growth \citep{dehghani2023,wortsman2024}; with the gain frozen at 1.0 the logits are capped at \(\sqrt{d}\) and no multiplicative logit scale can be learned.

\section{Setup}\label{setup}

\subsection{Models and training arms}\label{models-and-training-arms}

The Dense and Mixed twins are decoder-only transformers with RoPE (base $\theta$ = \(10^{6}\)), grouped-query attention with QK-normalisation, SwiGLU, RMSNorm, tied embeddings and no dropout, with a Qwen3 tokeniser (vocabulary 151,936; \citealp{yang2025}) and trained from scratch on 8,192-token FineWeb-Edu sequences (sample-100BT; \citealp{penedo2024}; Appendix A) with AdamW \citep{loshchilov2019} at $\beta_{1}$ = 0.9, $\beta_{2}$ = 0.95, peak learning rate 5$\times$\(10^{-4}\), weight decay 0.01, 750 warm-up steps and cosine decay. The tiers are 365M (hidden 768, 28 layers, 44,000 steps, batch 20 sequences, 7.2B tokens; seeds 42--47), 1.2B (1,536, 28 layers, 50,462 steps, batch 60, 24.8B tokens; seeds 42, 70--72) and 3B (2,304, 36 layers, 100,000 steps, batch 60, 49.2B tokens; seed 42); unique parameter counts are 359M, 1.20B and 3.00B. Twins share the seed, optimiser, schedule, step count and data order; only the training attention mask differs. The Dense arm uses full causal attention at every step. In the Mixed arm each attention layer, at every step, draws one causal block mask from a fixed pool of 10,000 (block size 128; shared across the batch, independent across layers) that retains \emph{k} of the 64 key blocks, \emph{k} uniform over \{63, 47, 31, 21, 11\} (budgets 98--17\%; Appendix A). At deployment a Mixed checkpoint is an ordinary dense transformer.

Three precision recipes are used (Table 3). \emph{Frozen gains}: the eleven original twin pairs were trained in pure bf16, every parameter stored and updated in bf16 with bf16 optimiser state and no fp32 master copy, and their normalisation gains stayed exactly at 1.0 (§4.2); the appendix calls this the canonical recipe. \emph{Learnable gains}: the same recipe with the 56 QK-normalisation gain tensors (the \emph{QK gains}) kept as fp32 master weights, cast to bf16 only in the forward pass; 365M seeds 42--44 and 1.2B seeds 70--72. \emph{Standard recipe}: fp32 master weights and fp32 AdamW state for every parameter, bf16 compute; 365M seeds 41--43. Four 365M control arms at five or six seeds each (a static fixed mask, a resampled fixed mask, attention dropout, a scalar RoPE-distance control) and five ALiBi twins (\citealp{press2022}; distance biases replace RoPE, everything else as Dense) were trained with the same pure-bf16 recipe as the frozen-gain pairs; two further ALiBi twins (seeds 42, 43) were trained with learnable gains and are paired with the learnable-gain Dense of the same seed. Two single seed-42 365M \emph{regime pairs} change one architectural setting with frozen gains (no QK-normalisation; RoPE base 10$^{4}$), and three single \emph{parameterisation ablations} vary which gains can learn (all 113 normalisation gains; one learnable scalar per Q/K norm with the vectors frozen; RoPE base 10$^{4}$ with learnable QK gains); a Dense model trained natively at 16k is a training-time control (Appendix A). Later models were evaluated on other clusters, where the evaluator reproduces the original results to within 0.04\% (Appendix B).

\subsection{Attention temperature and calibration}\label{attention-temperature-and-calibration}

The \emph{attention temperature} $\tau$ divides the logits, \(qk^{\top}/(\tau\sqrt{d})\), before the softmax; it is applied on the query side after QK-normalisation (which would otherwise absorb it), so $\tau$ \textless{} 1 sharpens attention. ``Calibrated'' means one $\tau$ (written $\tau^{*}$) per model and evaluation length, selected on calibration windows by summed negative log-likelihood over three corpora from a 13-point grid \{0.5, 0.6, 0.7, 0.75, 0.8, 0.85, 0.9, 0.95, 1.0, 1.1, 1.25, 1.5, 2.0\} and applied to disjoint held-out windows, with the same grid for both arms. Because the temperature is selected at the evaluation length, the calibrated comparison is not zero-shot; we use it to diagnose the zero-shot one. §4.5 tests a data-free rule that needs no selection. Moving both arms one grid step together changes the 30k gap by up to 0.03 log-PPL (Appendix C.9). We therefore treat 0.03 as the \emph{resolution} of the calibrated comparison: a residual inside it is not interpreted by size, only by the consistency of its sign across seeds and windows, and we call the arms ``level'' when that sign is not consistent. \emph{In-span} means positions or lengths at or below the 8k training length, and $\Delta$ denotes log PPL(change) $-$ log PPL(Dense), where the change is Mixed unless stated otherwise (the control arms and ALiBi in Table 2); $\Delta$~=~$-$0.10 means the changed model's perplexity is 9.5\% lower.

\subsection{Evaluation protocol and statistics}\label{evaluation-protocol-and-statistics}

Perplexity is measured under full attention on non-overlapping windows of the evaluation length from FineWeb-Edu (in-distribution validation), PG-19 \citep{rae2020} and arXiv. Four window sets are used (Appendix B); absolute perplexities are comparable only within one set.

\begin{table*}[tp]
\centering
\begingroup\scriptsize\setlength{\tabcolsep}{3pt}
\sbox{\tblbox}{\begin{tabular}{@{}lll@{}}
\toprule
Window set & Per corpus & Used for\tabularnewline
\midrule
Calibration & 8, 4, 2, 2 at 8k, 16k, 24k, 30k & selecting $\tau^{*}$\tabularnewline
Held-out & 8, 4, 2, 2, disjoint from calibration & every calibrated result; Tables 1--4 unless marked diagnostics\tabularnewline
36-window & 12 at 24k and 30k: the held-out two plus ten more & window-level intervals for single pairs\tabularnewline
Diagnostics & 24, 12, 8, 8 & default-temperature and position-resolved results of §4.1; Fig. 2\tabularnewline
\bottomrule
\end{tabular}}\ifdim\wd\tblbox>\linewidth\resizebox{\linewidth}{!}{\usebox{\tblbox}}\else\usebox{\tblbox}\fi
\endgroup
\caption*{\textbf{Window sets.} Windows per corpus at each evaluation length, and what each set is used for.}
\end{table*}

Position-resolved results use eight equal bins per window. Dense and changed runs sharing a seed are paired; the unit of analysis is the seed. For each comparison we report the mean paired $\Delta$, its SD across seeds and the number of seeds favouring the change; with three to six pairs the direction count carries the evidence, and we never pool tiers into one test. Protocol matching, provenance, the cross-hardware floor and the acceptance checks are in Appendix B.

\section{Results}\label{results}

\subsection{The default-temperature comparison and the calibrated one}\label{the-default-temperature-comparison-and-the-calibrated-one}

\begin{figure*}[tp]
\centering
\includegraphics[width=\textwidth]{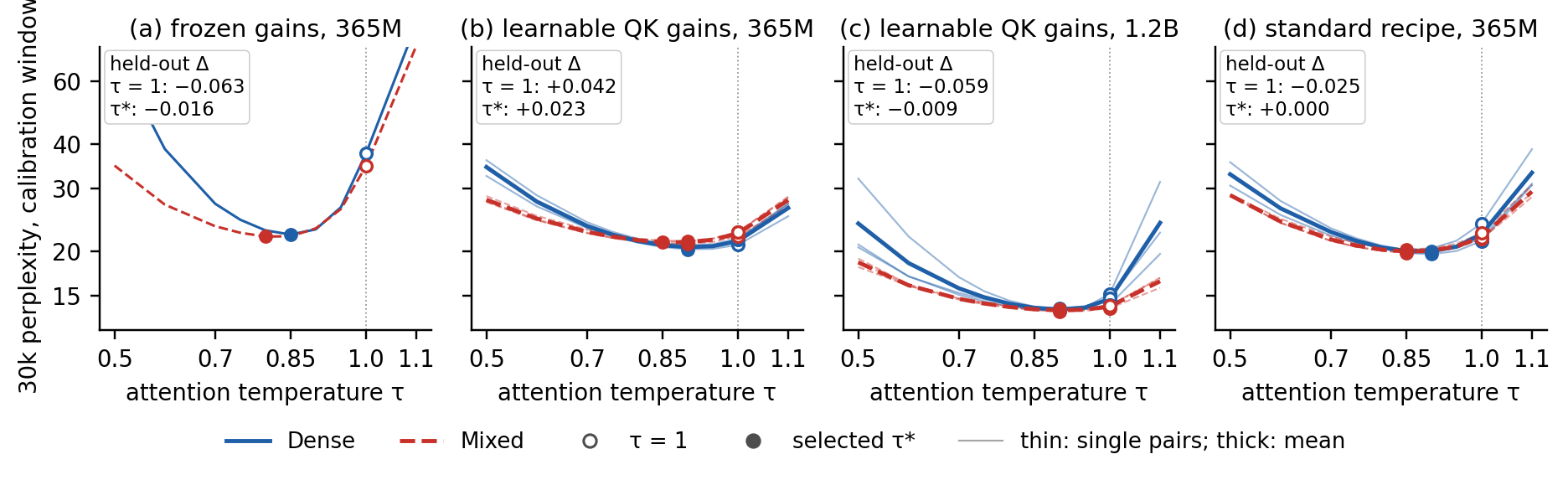}
\caption{\textbf{30k perplexity against the attention temperature for the Dense (solid) and Mixed (dashed) arms: (a) frozen gains, 365M seed 42; (b) learnable QK gains, 365M, three pairs; (c) learnable QK gains, 1.2B, three pairs; (d) the standard recipe, 365M, three pairs.} Curves: the 13-point sweep on the calibration windows, on which each arm's $\tau^{*}$ is selected (filled marker; open marker $\tau$~=~1; thin lines single pairs, thick their mean); these windows are easier than the held-out windows quoted in the text, so the curves sit lower than those numbers. Boxes: the held-out $\Delta$~=~log PPL(Mixed) $-$ log PPL(Dense) at $\tau$~=~1 and at $\tau^{*}$, seed-averaged (Table 1 for (a); Table 3 for (b)--(d)). With frozen gains (a) the baseline sits far up the right-hand branch at $\tau$~=~1 while Mixed's curve is flatter there; the later recipes shrink that climb without removing it.}
\end{figure*}

\begin{figure*}[tp]
\centering
\includegraphics[width=0.70\textwidth]{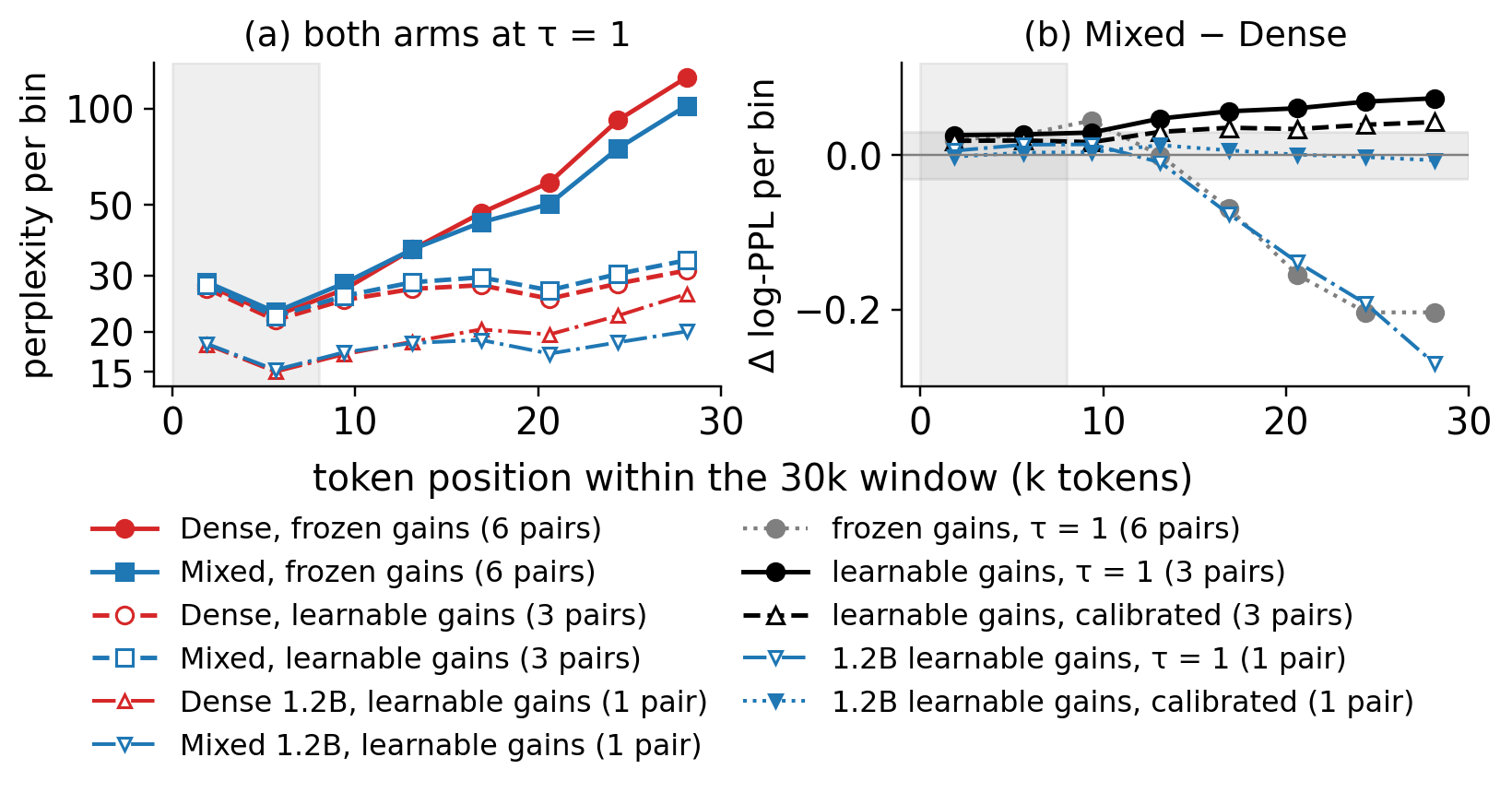}
\caption{\textbf{Position-resolved perplexity at 30k with frozen and with learnable QK-normalisation gains (365M, diagnostics windows, eight bins, geometric mean over corpora).} (a) Both arms at $\tau$~=~1: with frozen gains (six pairs; solid) both arms climb to 100--125 by the last bin and Mixed climbs less; with learnable gains (three pairs; dashed) both stay near 30 and Mixed is slightly worse throughout. (b) Mixed~$-$~Dense per bin: the frozen-gain default advantage grows to $-$0.20 (grey dotted); with learnable gains Mixed is worse in every bin at $\tau$~=~1 (solid) and calibrated (dashed), in all three seeds. Dash-dotted: the 1.2B seed-70 pair with learnable gains, whose Dense arm still rises from 18 to 26 at $\tau$~=~1 while Mixed rises only to 20, so Mixed's $\tau$~=~1 advantage grows to $-$0.27 in the last bin; seeds 71 and 72 are in Appendix C.4.}
\end{figure*}

Figure 1 plots each arm's 30k perplexity against the temperature under the three recipes; this section quantifies panel (a), §4.3 panels (b, c) and §4.4 panel (d). With frozen gains the dense baseline is heavily miscalibrated beyond its training length. Its 30k held-out perplexity at $\tau$~=~1 is 55.9 at 365M (seed 42), 47.8--56.1 at 1.2B and 47.2 at 3B; at its own $\tau^{*}$ (0.85 in every case) it is 35.3, 23.3--24.0 and 20.2, a miscalibration of 37\% at 365M and 50--58\% at 1.2B and 3B. Binned by position, its $\tau$~=~1 perplexity rises 4.5$\times$ across a 30k window at 365M (Fig. 2a, solid). Mixed degrades less at $\tau$~=~1, so at 30k it looks 6--31\% better than Dense (Tables 1--2), and the advantage grows with length and with position in the window. On the diagnostics windows, which are a different sample from the held-out windows of Tables 1--2 and so give slightly different values, $\Delta$ goes from +0.026 (8k) to $-$0.067 (30k) at 365M (six pairs, all six), from +0.018 to $-$0.198 at 1.2B (three pairs) and from $-$0.001 to $-$0.343 at 3B (Appendix C.1); in the last position bin the advantage reaches $-$0.37 on FineWeb-Edu at 365M, $-$0.85 on FineWeb-Edu at 1.2B and $-$1.19 on PG-19 at 3B (Appendix C.3). The control arms show the same pattern in both directions (Table 2): attention dropout looks 8\% better than Dense at 30k (5 of 5), the five ALiBi twins 35\% better (5 of 5), the two fixed-mask arms 30--36\% worse and the RoPE-distance control 5--6\% worse.

\begin{table*}[tp]
\centering
\begingroup\footnotesize\setlength{\tabcolsep}{4pt}
\sbox{\tblbox}{\begin{tabular}{@{}llrrll@{}}
\toprule
Tier & Seed & \shortstack[l]{$\Delta$\\ default} & \shortstack[l]{$\Delta$\\ calibrated} & \shortstack[l]{36-window\\ $\Delta\pm$ 95\%} & \shortstack[l]{Mixed better\\ (of 36)}\tabularnewline
\midrule
365M & 42 & $-$0.063 & $-$0.016 & $-$0.014 $\pm$ 0.004 & 31/36\tabularnewline
1.2B & 42 & $-$0.147 & $-$0.002 & +0.001 $\pm$ 0.005 & 18/36\tabularnewline
1.2B & 70 & $-$0.378 & $-$0.004 & $-$0.006 $\pm$ 0.005 & 25/36\tabularnewline
1.2B & 71 & $-$0.239 & +0.001 & +0.006 $\pm$ 0.005 & 15/36\tabularnewline
1.2B & 72 & $-$0.171 & +0.025 & +0.021 $\pm$ 0.008 & 9/36\tabularnewline
3B & 42 & $-$0.368 & +0.003 & +0.010 $\pm$ 0.014 & 20/36\tabularnewline
\bottomrule
\end{tabular}}\ifdim\wd\tblbox>\linewidth\resizebox{\linewidth}{!}{\usebox{\tblbox}}\else\usebox{\tblbox}\fi
\endgroup
\caption*{\textbf{Table 1:} Calibrated comparison at 30k for the six frozen-gain pairs with their own temperature selection (365M seed 42, the four 1.2B pairs, 3B). $\Delta$: corpus-mean log PPL(Mixed) $-$ log PPL(Dense) on the held-out windows. Last two columns: the same pair on the 36-window set, mean $\pm$ 95\% half-width and windows with Mixed ahead; these describe one pair, not seeds. The 8k rows: Appendix C.1.}
\end{table*}

\begin{table*}[tp]
\centering
\begingroup\scriptsize\setlength{\tabcolsep}{3pt}
\sbox{\tblbox}{\begin{tabular}{@{}llllll@{}}
\toprule
\shortstack[l]{Change (vs\\ its Dense twin)} & n & \shortstack[l]{$\Delta$ at\\ $\tau$~=~1} & \shortstack[l]{Better ($\tau$\\ = 1)} & \shortstack[l]{$\Delta$\\ calibrated} & \shortstack[l]{Better\\ ($\tau^{*}$)}\tabularnewline
\midrule
Mixed, 365M (a) & 6 & $-$0.067 & 6/6 & +0.007 & 2/6\tabularnewline
Mixed, 1.2B & 4 & $-$0.234 & 4/4 & +0.005 & 2/4\tabularnewline
Mixed, 3B & 1 & $-$0.368 & 1/1 & +0.003 & 0/1\tabularnewline
Static fixed mask, 365M & 6 & +0.307 & 0/6 & $-$0.007 & 4/6\tabularnewline
Resampled fixed mask, 365M & 6 & +0.262 & 0/6 & $-$0.008 & 5/6\tabularnewline
Attention dropout, 365M & 5 & $-$0.085 & 5/5 & $-$0.009 & 4/5\tabularnewline
RoPE-distance control, 365M & 5 & +0.054 & 0/5 & +0.001 & 1/5\tabularnewline
ALiBi vs RoPE, 365M & 5 & $-$0.427 & 5/5 & +0.046 & 0/5\tabularnewline
NTK scaling (positive control) (b) & 3 & $-$0.473 to $-$0.083 & 3/3 & $-$0.046 to $-$0.034 & 3/3\tabularnewline
Mixed, 365M, learnable gains & 3 & +0.042 & 0/3 & +0.023 & 0/3\tabularnewline
Mixed, 1.2B, learnable gains & 3 & $-$0.059 & 3/3 & $-$0.009 & 3/3\tabularnewline
Mixed, 365M, standard recipe & 3 & $-$0.025 & 3/3 & +0.000 & 1/3\tabularnewline
ALiBi vs RoPE, 365M, learnable gains & 2 & +0.086 & 0/2 & +0.120 & 0/2\tabularnewline
Mixed, 365M, all 113 gains fp32 & 1 & +0.034 & 0/1 & +0.023 & 0/1\tabularnewline
Mixed, 365M, scalar Q/K gains & 1 & +0.005 & 0/1 & $-$0.005 & 1/1\tabularnewline
Dense-16k vs Dense-8k, learnable gains (control) & 1 & +0.004 & 0/1 & +0.028 & 0/1\tabularnewline
Mixed, 365M, no QK-norm & 1 & +0.127 & 0/1 & +0.197 & 0/1\tabularnewline
Mixed, 365M, RoPE base 10$^{4}$ & 1 & $-$0.051 & 1/1 & $-$0.051 & 1/1\tabularnewline
Mixed, 365M, RoPE base 10$^{4}$, learnable gains & 1 & +0.016 & 0/1 & +0.013 & 0/1\tabularnewline
\bottomrule
\end{tabular}}\ifdim\wd\tblbox>\linewidth\resizebox{\linewidth}{!}{\usebox{\tblbox}}\else\usebox{\tblbox}\fi
\endgroup
\caption*{\textbf{Table 2:} Every training-time change through the protocol at 30k. $\Delta$~=~log PPL(change) $-$ log PPL(its Dense twin), averaged over pairs (negative favours the change); ``Better'' counts pairs with $\Delta$ \textless{} 0; corpus means on the held-out windows (the four in-house control arms and frozen-gain ALiBi: per-seed log ratios of pooled held-out perplexity). Rows not marked otherwise are frozen-gain models. (a) Six seeds on the diagnostics windows at $\tau$~=~1; seed 42's temperatures applied to all six at $\tau^{*}$. (b) Inference-time positive control on Dense seed 42 of each tier; range over tiers. Per-pair values for the twin pairs: Appendix C.}
\end{table*}

Calibration removes all of it. For the six pairs with their own temperature selection, the calibrated 30k gap lies between $-$0.016 and +0.025 (Table 1): $-$0.368 becomes +0.003 at 3B, and the four 1.2B gaps of $-$0.147 to $-$0.378 become $-$0.004 to +0.025. For the other five 365M pairs we apply seed 42's temperatures; the six-seed mean is then +0.007, with Mixed better in 2 of 6 (Appendix C.1). The 36-window residuals are $-$0.014 to +0.021 with half-widths of 0.004--0.014 (Table 1), and the seed-level 90\% equivalence intervals, {[}$-$0.004, +0.019{]} at 365M (at seed 42's transferred temperatures) and {[}$-$0.011, +0.021{]} at 1.2B, lie within the $\pm$0.03 resolution (Appendix C.2). At $\tau^{*}$ the Dense per-bin perplexity no longer rises with position at 365M and 1.2B (4.6\% in the last bin at 3B), and the per-bin Mixed advantage stays within $\pm$0.02 at 365M (six-seed mean at seed 42's temperatures; seed 42 at its own $\tau^{*}$ reaches $-$0.046 in one bin) and $\pm$0.017 at 1.2B (Appendix C.3). The control arms' differences go the same way (Table 2): dropout's 8\% becomes 1\%, the fixed masks' 30--36\% becomes under 1\%, and ALiBi's 35\% advantage becomes a 5\% deficit (0 of 5), close to its 6\% in-span deficit. Almost all of the ALiBi change comes from calibrating the RoPE baseline, since ALiBi's own perplexity is nearly flat in $\tau$ ($\tau^{*}$~=~0.95 at every length; Appendix C.6). The positive control survives: NTK-aware RoPE scaling at inference gains 8--38\% at 30k at $\tau$~=~1 and still 3--4.5\% once both models are calibrated, at every tier and extension length (Appendix C.9). With frozen gains the default comparison misreports every training-time change in this section, by up to 0.37 log-PPL for Mixed and 0.49 for ALiBi.

\subsection{The frozen normalisation gains}\label{the-frozen-normalisation-gains}

The frozen-gain baseline of §4.1 is miscalibrated because it could not learn a logit scale at all. In all 22 frozen-gain checkpoints every stored QK-normalisation gain value is exactly 1.0, and in the Dense-42 checkpoint all 113 normalisation-gain tensors are, the residual RMSNorm gains included (Appendix B). The gains are ordinary trainable AdamW parameters, so they receive an update at every step, but each update rounds back to 1.0. The gains were stored and updated in bf16, whose spacing at 1.0 is \(2^{-7}\) above and \(2^{-8}\) below, so under round-to-nearest an update smaller than \(2^{-9} \approx 1.95 \times 10^{-3}\) in magnitude leaves 1.0 unchanged. An AdamW update is \(\eta\,\hat m/(\sqrt{\hat v}+\epsilon)\) with bias-corrected moments, and by Cauchy--Schwarz \(|\hat m|/\sqrt{\hat v} \le (1-\beta_1)/\sqrt{(1-\beta_2)(1-\beta_1^2/\beta_2)}\), which is 1.165 at $\beta_{1}$ = 0.9, $\beta_{2}$ = 0.95 (7.27 at $\beta_{2}$ = 0.999; Appendix B). A gain at 1.0 therefore cannot move at any peak learning rate below \(2^{-9}/1.165 \approx 1.7 \times 10^{-3}\) (2.7$\times$\(10^{-4}\) at $\beta_{2}$ = 0.999); at our peak rate of 5$\times$\(10^{-4}\) the largest possible step is 5.8$\times$\(10^{-4}\). fp32 master weights for the gains, or stochastic rounding, prevent the freeze at any learning rate, and the learnable-gain and standard-recipe checkpoints of §4.3--4.4 have 99\% of their gain values more than one bf16 spacing away from 1.0. With unit gains after QK-normalisation the logits are capped at \(\sqrt{d}\), so a frozen-gain model cannot learn a multiplicative attention logit scale, the quantity a temperature corrects. Within the training span the freeze costs 2--5\% in perplexity at 365M (Appendix C.4); at 1.2B the in-span difference is under 2\% and not consistent in sign for Dense (Table 3). Beyond it the freeze enlarges the miscalibration of §4.1: at 365M it is 37--39\% with frozen gains and 2--4\% once the QK gains can learn (§4.3).

\subsection{Unfreezing the gains}\label{unfreezing-the-gains}

\textbf{At 365M} (seeds 42--44; Table 4) the gains moved. The mean QK gain is 1.31 in Dense and 1.41 in Mixed, and 99\% of the values differ from 1.0 by more than the bf16 spacing. The mean product of query and key gains is 1.8 (Dense) and 2.1 (Mixed) times the frozen one and is frequency-selective, so the freeze prevented a per-dimension re-weighting as well as a scale (Appendix C.11). Dense's 30k perplexity at $\tau$~=~1 drops from 56--57 to 33--34, its miscalibration from 37--39\% to 2--4\%, and its rise across a 30k window from 4.5$\times$ to 14\% (Fig. 2a, dashed). Mixed is now behind at 30k at both temperatures (+0.042 at $\tau$~=~1, +0.023 calibrated, 3 of 3 seeds, 36 of 36 windows in every seed, every position bin; Fig. 2b) and behind in-span too (+0.024 at 8k). At 8k the learnable-gain Dense is 2.6--4.6\% better than its frozen twin.

\begin{table*}[tp]
\centering
\begingroup\scriptsize\setlength{\tabcolsep}{3pt}
\sbox{\tblbox}{\begin{tabular}{@{}llllll@{}}
\toprule
Recipe & \shortstack[l]{Tier\\ (pairs)} & \shortstack[l]{Dense\\ PPL 8k} & \shortstack[l]{Dense PPL 30k $\tau$\\ = 1 $\rightarrow\tau$* (miscalibration)} & \shortstack[l]{$\Delta$ 30k, $\tau$\\ = 1 (ahead)} & \shortstack[l]{$\Delta$ 30k,\\ $\tau^{*}$ (ahead)}\tabularnewline
\midrule
Frozen gains & 365M (6) & 32.8 & 55.9 $\rightarrow$ 35.3 (37\%) & $-$0.067 (6/6) & +0.007 (2/6)\tabularnewline
Frozen gains & 1.2B (4) & 21.7--22.1 & 47.8--56.1 $\rightarrow$ 23.3--24.0 (50--58\%) & $-$0.234 (4/4) & +0.005 (2/4)\tabularnewline
Frozen gains & 3B (1) & 18.6 & 47.2 $\rightarrow$ 20.2 (57\%) & $-$0.368 (1/1) & +0.003 (0/1)\tabularnewline
Learnable QK gains (56 fp32 tensors) & 365M (3) & 31.6 & 34.0 $\rightarrow$ 32.8 (3\%) & +0.042 (0/3) & +0.023 (0/3)\tabularnewline
Learnable QK gains (56 fp32 tensors) & 1.2B (3) & 21.7 & 24.3 $\rightarrow$ 22.6 (7\%) & $-$0.059 (3/3) & $-$0.009 (3/3)\tabularnewline
All 113 normalisation gains fp32 & 365M (1) & 30.5 & 33.8 $\rightarrow$ 31.8 (6\%) & +0.034 (0/1) & +0.023 (0/1)\tabularnewline
Standard recipe (all parameters fp32 master) & 365M (3) & 29.2--29.7 & 34.5--37.9 $\rightarrow$ 31.6--32.3 (8--15\%) & $-$0.025 (3/3) & +0.000 (1/3)\tabularnewline
\bottomrule
\end{tabular}}\ifdim\wd\tblbox>\linewidth\resizebox{\linewidth}{!}{\usebox{\tblbox}}\else\usebox{\tblbox}\fi
\endgroup
\caption*{\textbf{Table 3:} The dense baseline's miscalibration at 30k and the Mixed~$-$~Dense gap, by precision recipe. PPL: pooled held-out perplexity, geometric mean over the three corpora (frozen 365M is seed 42, the only frozen-gain 365M pair with its own selection; other rows are ranges or means over the pairs). Miscalibration: Dense's perplexity drop from $\tau$~=~1 to $\tau^{*}$. $\Delta$: log PPL(Mixed) $-$ log PPL(Dense); ``ahead'' counts pairs with $\Delta$ \textless{} 0 (frozen 365M: six seeds on the diagnostics windows at $\tau$~=~1, seed 42's temperatures applied to all six at $\tau^{*}$). Per seed: Table 4 and Appendix C.1, C.4, C.5, C.8.}
\end{table*}

\begin{table*}[tp]
\centering
\begingroup\scriptsize\setlength{\tabcolsep}{3pt}
\sbox{\tblbox}{\begin{tabular}{@{}llllrrllr@{}}
\toprule
Recipe & Seed & \shortstack[l]{$\tau^{*}$ D\\ / M} & \shortstack[l]{Dense PPL $\tau$\\ = 1 $\rightarrow\tau$*} & \shortstack[l]{$\Delta\tau$\\ = 1} & $\Delta\tau$* & \shortstack[l]{36w $\Delta\tau$\\ = 1 (ahead)} & \shortstack[l]{36w\\ $\Delta\tau$* (ahead)} & \shortstack[l]{$\Delta$\\ 8k}\tabularnewline
\midrule
Learnable QK gains, 365M & 42 & 0.90 / 0.90 & 33.1 $\rightarrow$ 32.3 (2\%) & +0.051 & +0.029 & +0.056 $\pm$ 0.006 (0/36) & +0.034 $\pm$ 0.005 (0/36) & +0.032\tabularnewline
Learnable QK gains, 365M & 43 & 0.90 / 0.85 & 34.5 $\rightarrow$ 33.0 (4\%) & +0.037 & +0.024 & +0.043 $\pm$ 0.007 (0/36) & +0.028 $\pm$ 0.005 (0/36) & +0.022\tabularnewline
Learnable QK gains, 365M & 44 & 0.90 / 0.90 & 34.4 $\rightarrow$ 33.2 (3\%) & +0.037 & +0.017 & +0.045 $\pm$ 0.005 (0/36) & +0.024 $\pm$ 0.004 (0/36) & +0.019\tabularnewline
Learnable QK gains, 1.2B & 70 & 0.90 / 0.90 & 25.2 $\rightarrow$ 22.4 (11\%) & $-$0.091 & $-$0.001 & $-$0.077 $\pm$ 0.008 (36/36) & +0.005 $\pm$ 0.004 (10/36) & +0.006\tabularnewline
Learnable QK gains, 1.2B & 71 & 0.90 / 0.90 & 24.3 $\rightarrow$ 22.7 (7\%) & $-$0.068 & $-$0.016 & $-$0.058 $\pm$ 0.011 (36/36) & $-$0.015 $\pm$ 0.007 (24/36) & +0.004\tabularnewline
Learnable QK gains, 1.2B & 72 & 0.90 / 0.90 & 23.5 $\rightarrow$ 22.7 (3\%) & $-$0.019 & $-$0.010 & $-$0.011 $\pm$ 0.008 (22/36) & $-$0.004 $\pm$ 0.005 (22/36) & +0.010\tabularnewline
Standard recipe, 365M & 41 & 0.90 / 0.85 & 34.6 $\rightarrow$ 31.6 (9\%) & $-$0.004 & +0.003 & $-$0.005 $\pm$ 0.007 (18/36) & $-$0.001 $\pm$ 0.005 (19/36) & +0.031\tabularnewline
Standard recipe, 365M & 42 & 0.90 / 0.85 & 34.5 $\rightarrow$ 31.7 (8\%) & $-$0.002 & $-$0.003 & +0.005 $\pm$ 0.007 (20/36) & +0.002 $\pm$ 0.003 (17/36) & +0.021\tabularnewline
Standard recipe, 365M & 43 & 0.85 / 0.85 & 37.9 $\rightarrow$ 32.3 (15\%) & $-$0.070 & +0.001 & $-$0.062 $\pm$ 0.016 (31/36) & +0.010 $\pm$ 0.011 (20/36) & +0.020\tabularnewline
\bottomrule
\end{tabular}}\ifdim\wd\tblbox>\linewidth\resizebox{\linewidth}{!}{\usebox{\tblbox}}\else\usebox{\tblbox}\fi
\endgroup
\caption*{\textbf{Table 4:} The three recipes with three seed pairs, per seed, at 30k. $\tau^{*}$: selected temperature (Dense / Mixed). Dense PPL: pooled held-out perplexity at $\tau$~=~1 $\rightarrow\tau$* (miscalibration in brackets). $\Delta$: corpus-mean log PPL(Mixed) $-$ log PPL(Dense) on the held-out windows at $\tau$~=~1 and at $\tau^{*}$; 36w: the same on the 36-window set, mean $\pm$ 95\% half-width and windows with Mixed ahead. $\Delta$ 8k: in-span gap at $\tau$~=~1.}
\end{table*}

\textbf{At 1.2B} (seeds 70--72; Table 4) Dense's 30k perplexity at $\tau$~=~1 is 23.5--25.2 instead of 48--56 and its miscalibration 3--11\% instead of 52--58\%. The default 30k gap shrinks by 71--89\% (mean 79\%), with Mixed still ahead at $\tau$~=~1 in 36 of 36 windows for seeds 70 and 71. Calibration puts the remainder within the resolution in all three seeds. Mixed is ahead in 3 of 3 seeds but in only 10--24 of 36 windows, so we read the arms as level and not as a Mixed advantage. This differs from 365M, where Mixed is behind. The in-span gap is +0.004 to +0.010.

\textbf{ALiBi.} An ALiBi arm trained with learnable QK gains (seeds 42, 43), paired with the learnable-gain Dense of the same seed, is behind RoPE at 30k even at $\tau$~=~1 (+0.113 and +0.059) and further behind calibrated (+0.137 and +0.103; ahead in 0 and 8 of 36 windows). Its own temperature response is flat ($\tau^{*}$~=~0.95 and 1.00 at 30k, miscalibration 0.0--0.1\%) and its in-span deficit is 4--6\%, close to its deficit with frozen gains (Appendix C.6). So the 35\% frozen-gain ALiBi advantage came from the miscalibrated RoPE baseline, and in this setup ALiBi extrapolates no better than RoPE.

\textbf{Which part of the gain matters.} Two seed-42 ablations at 365M separate the scalar logit scale from the per-dimension re-weighting (Appendix C.8). With one learnable scalar per Q/K norm and the gain vectors frozen, Dense's miscalibration is gone ($\tau^{*}$~=~0.95 at 24k--30k, 0.1--0.4\%) and Mixed is level with Dense at every length ($\Delta$ within $\pm$0.008). The scalar-gain Dense is 3.5\% worse in-span than the learnable-gain Dense, whereas the scalar-gain Mixed is not worse than its learnable-gain counterpart; the per-dimension re-weighting is therefore worth a few per cent to Dense and nothing to Mixed (one seed). Unfreezing all 113 normalisation gains leaves the comparison unchanged (+0.034 at $\tau$~=~1, +0.023 calibrated) and improves both arms by 2.7--3.7\% in-span. Two further single pairs test the architectural dependence (Table 2; Appendix C.8). Without QK-normalisation both arms' 30k perplexity at $\tau$~=~1 is about twice the frozen-gain value (Dense 108) and calibration widens Dense's lead to +0.20, so the freeze explains the size of the Mixed--Dense gap but not all of Dense's miscalibration. At RoPE base 10$^{4}$ a 5\% calibrated Mixed advantage survives with frozen gains ($-$0.051; 26 and 36 of 36 windows at $\tau$~=~1 and $\tau^{*}$). It is the only calibrated Mixed advantage beyond the resolution in any recipe at 30k, and it disappears once the gains can learn (+0.016 and +0.013).

\subsection{The standard recipe}\label{the-standard-recipe}

The standard-recipe pairs (seeds 41--43; Table 4) test whether the confound needs pure-bf16 training. In-span they are the best 365M models in the study: Dense's 8k perplexity is 29.2--29.7 against 31.3--31.8 with learnable QK gains and 32.7--32.8 with frozen gains at the same seeds as the learnable-gain pairs (42--44). The 57 residual and final-norm gains, frozen in the other two recipes, settle at 0.86--0.89 on average, and the QK gains land where the learnable-gain recipe put them (Dense 1.28--1.30, Mixed 1.39--1.40; Appendix C.11). Beyond the span the baseline is more miscalibrated than with learnable QK gains alone: 8.6\%, 8.2\% and 14.7\% at 30k (Table 4, rounded there), against 2--4\% with learnable QK gains and 37--39\% frozen. The miscalibration therefore does not require the freeze, although frozen gains make it several times larger.

In two of the three pairs Mixed and Dense are level at 30k at both temperatures (Table 4). In the third, seed 43, the baseline's miscalibration is the largest of the three and Mixed is 7\% better at $\tau$~=~1 (ahead in 31 of 36 windows) and level at $\tau^{*}$ (20 of 36): the full confound, under the standard recipe, in one seed of three. In all three pairs Mixed is 2--3\% worse in-span and level or behind calibrated at every length (Appendix C.5). Across recipes, Mixed is 2--3\% worse than Dense in-span in the six three-seed 365M pairs, within about 1\% in the 1.2B learnable-gain pairs, the 3B pair and the scalar-gain pair, and 0.5--4\% worse in the remaining pairs. Once both arms are calibrated it is level or behind in every pair whose gains could learn (nine at 365M, three at 1.2B). Stochastic block-sparse pretraining does not extend context at any precision we tried. Its default-temperature advantage comes from the baseline's miscalibration and varies with recipe and scale: 2.5\% on average (7\% in one seed) under the standard recipe at 365M, 6.5\% with frozen gains at 365M and 31\% with frozen gains at 3B.

\subsection{The check}\label{the-check}

The check has four parts. For each model and evaluation length, select one attention temperature on calibration windows and score held-out windows at it, with the same grid for every arm. Report the calibrated comparison next to the default one. State the resolution (here 0.03 log-PPL, one grid step) and read residuals inside it by sign consistency, not by size. Include a positive control, a change whose gain should survive, to show that the procedure does not also remove real gains; NTK-aware scaling plays that role here, with a calibrated gain of 3--4.5\% at every tier and length on the frozen-gain Dense checkpoints. Applied without knowledge of the cause, the procedure closes the frozen-gain gaps (the RoPE-10$^{4}$ gap needs learnable gains as well) and exposes the seed-43 standard-recipe gain. With no per-arm search, the entropy-invariance rule $\tau$~=~ln(8,192)/ln(L) \citep{su2021} applied to both arms lands within one grid step of the Dense selection and leaves \textbar $\Delta$\textbar{} $\leq$ 0.023 at one pair per tier, whereas YaRN's sharper rule leaves Mixed 2--6\% ahead (Appendix C.10). The check says nothing about which arm is better in-span or whether either arm uses the far context. After calibration the per-bin loss shows no consistent decline beyond the first bins (Appendix C.3), and perplexity is an insensitive measure of long-range use \citep{fang2025}, the more so on windows that cross document boundaries (Limitations).

\section{Recommendations and conclusion}\label{recommendations-and-conclusion}

A training-time attention change evaluated beyond its training length should be compared with its baseline at calibrated temperatures as well as at the default, with a positive control and a stated resolution. This holds under the standard recipe: in our three standard-recipe pairs the 8k baseline is 8--15\% miscalibrated at 30k, and in one of them this gave the changed model a 7\% default-temperature advantage that calibration removed. The check costs one temperature sweep per model and evaluation length, or none with the entropy-invariance rule; for other changes the grid step and the positive control have to be chosen again. The second check is whether the normalisation gains moved. Under round-to-nearest bf16 without fp32 master weights they cannot move at any peak learning rate below about 1.7$\times$10$^{-3}$ with $\beta_{2}$ = 0.95 (2.7$\times$\(10^{-4}\) with $\beta_{2}$ = 0.999), and a one-line read of the checkpoint shows whether they did. A baseline with frozen gains has a 30k perplexity at $\tau$~=~1 of 1.6--2.4 times its calibrated value, against which a sparse, dropout or ALiBi variant that tolerates the default temperature better appears to extrapolate; fp32 master weights \citep{micikevicius2018} for the gains remove most of the effect.

\section*{Acknowledgements}
The author thanks his supervisor, Martin Jaggi, for hosting this work at the Machine Learning and Optimization Laboratory at EPFL, which provided the compute for the models and evaluations reported here.

\section*{Limitations}
\textbf{Seed counts and scale.} Six paired seeds at 365M and four pairs at 1.2B carry the frozen-gain result; the 3B pair, the regime pairs, the parameterisation ablations and the Dense-16k control are single pairs. The learnable-gain result has three pairs at each of 365M and 1.2B and the standard-recipe result three pairs at 365M only; the 3B twins were not retrained, so the growth of the frozen-gain effect with scale (§4.1) is documented under the frozen-gain recipe only, and the standard-recipe miscalibration is measured at one scale. Effects that appear larger at 1.2B and 3B conflate parameters with training tokens (7.2B/24.8B/49.2B) and with data repetition (1.98 passes at 3B, 1.00 at 1.2B, 0.29 at 365M). The ALiBi comparison has five frozen-gain pairs and two learnable-gain pairs; the four in-house control arms were run with frozen gains only.

\textbf{Calibration.} Among the frozen-gain twins, temperature selection covers one seed at 365M and 3B and four at 1.2B; the six-seed 365M check applies seed 42's temperatures without re-selecting them and is directional support only. The window-level intervals describe one pair on 36 windows, do not support seed-level inference, and treat windows cut from one token stream as independent, so autocorrelation would make them anti-conservative. The $\pm$0.03 resolution is applied to every calibrated residual; residuals inside it are read from their sign consistency across seeds and windows, which for the 1.2B learnable-gain pairs favours Mixed in 3 of 3 seeds but in only 10--24 of 36 windows, and for the 365M learnable-gain and standard-recipe pairs is against Mixed or mixed. The NTK positive control's calibrated gain is one to one and a half grid steps, so the procedure's ability to preserve a genuine gain is shown near its own resolution, and only on frozen-gain baselines: the control was not run on the learnable-gain or standard-recipe checkpoints. We did not tune per-layer, per-head, per-position or learned temperatures. ALiBi's temperature rescales its content logits while its distance biases stay fixed, so the ALiBi comparison tests the RoPE baseline's calibration rather than ALiBi's own; a temperature on the full ALiBi score was not tried.

\textbf{Breadth.} The study uses a single architecture family (a Qwen3-style configuration with RoPE base \(10^{6}\) except for the RoPE-10$^{4}$ regime pair and ablation, a setting where softmax flattening with key count is expected and positional failure is mild), a single pretraining corpus, one training length for the twin pairs and one mask family. The two regime pairs show that the calibrated result changes with QK-normalisation and RoPE base, but they are single seeds, and the no-QK-norm pair selects the sharpest temperatures in the paper (0.70--0.75). Whether the effect persists with native long-context training, other data or other architectures is untested; the Dense-16k control (Table 2; Appendix C.8), trained on its 8k twin's tokens in half the steps, did not improve on the twin and is not a clean positive control. Evaluation windows are contiguous slices of a concatenated token stream and can cross document boundaries; on FineWeb-Edu most of a 30k window is unrelated preceding text, so position-resolved results there may not measure long-range use, and we did not score the same targets with the distant context removed. Why the frozen scale costs Dense more than Mixed, and why the standard-recipe Dense is more miscalibrated beyond the span than the learnable-gain Dense, are not identified. The evidence comes from several evaluation runs on three clusters. Protocol keying and the cross-hardware floor (0.04\% in perplexity, 0.0004 nats) are documented in Appendix B, but code revisions and example hashes were not recorded for every early result, the later ablation and control runs were made from an uncommitted working copy of the training code, two frozen-gain 1.2B arms are repaired runs restarted from an earlier checkpoint, and both arms of the 1.2B learnable-gain pairs were trained across interrupted launches resumed from checkpoints.

\section*{Ethical considerations}
The models are small base language models trained on a public web corpus (FineWeb-Edu) and evaluated on public corpora; no human subjects or annotation were involved and no personal data were collected (the public corpora may contain personal information present in their sources), and no model is released with capabilities beyond those of comparably sized public checkpoints. The study's main cost is training compute: 25 Dense/Mixed pairs up to 3B parameters (six learnable-gain, three standard-recipe, two regime pairs and three parameterisation ablations), plus 29 control-arm models (seven of them ALiBi twins) and one Dense-16k model at 365M, trained data-parallel on A100 and H100 GPUs in allocations that varied by run, on three clusters. Per-run GPU-hours were not recorded by a scheduler; a 6ND estimate with the unique parameter counts gives \(5.3 \times 10^{21}\) training FLOPs for the models reported here (pilots, superseded and incomplete runs excluded), roughly 3,000--4,900 H100-hours or 9,400--15,600 A100-hours at 30--50\% utilisation. The training and evaluation code is at \url{https://github.com/rubentium/SparseAtt}; model checkpoints are not released.

\bibliography{refs}

\clearpage
\appendix
\setcounter{figure}{0}
\renewcommand{\thefigure}{A\arabic{figure}}
\section{Training details}\label{training-details}

\textbf{Architecture.} 365M: 359.0M unique parameters, hidden 768, 28 layers, 6 query and 2 KV heads, head dimension 128, MLP width 3,072. 1.2B: 1,202.4M parameters, hidden 1,536, 28 layers, 12 query and 4 KV heads, head dimension 128, MLP width 6,144. 3B: 3,004.4M parameters, hidden 2,304, 36 layers, 24 query and 8 KV heads, head dimension 128, MLP width 7,936. The input embedding and the output projection are tied (\path{tie_word_embeddings: true}; the 151,936 $\times$ \emph{d} embedding is counted once). Normalisation layers compute \(\hat{x} = x/\sqrt{\mathrm{mean}(x^2) + \epsilon}\) and return \(\hat{x} \cdot w\) with \emph{w} initialised to ones (not a \(1 + w\) parameterisation). RoPE uses the rotate-half convention with base \(10^{6}\) (10$^{4}$ in the base-10$^{4}$ regime pair).

\textbf{Optimisation.} Fused AdamW with $\beta_{1}$ = 0.9, $\beta_{2}$ = 0.95 and $\epsilon$ = 10$^{-8}$, peak learning rate 5$\times$\(10^{-4}\), linear warm-up over 750 steps then cosine decay to zero, weight decay 0.01 applied to every parameter (no exclusion of biases or normalisation weights), gradient clipping 1.0. Global batch 20 sequences at 365M and 60 at 1.2B and 3B, held fixed by gradient accumulation when the number of GPUs changed. A lower learning rate (3$\times$\(10^{-4}\)) was held in reserve for unstable runs; no run is recorded as having used it. Training is data-parallel in PyTorch \citep{paszke2019}; the Mixed arm uses FlexAttention \citep{dong2024} block masks on the compiled block-sparse path. Twins sharing a seed are configured to see the same training sequences in the same order: the training code resets the data RNG so that the data order is fixed by the seed, and the mask sampler draws from a separate stream. Runs were checkpointed and resumed whenever the allocation changed, with optimiser, scheduler, data-loader and RNG state restored and the scheduled learning rate preserved; the serialised data trajectories of resumed runs were not re-read, so identical example order across resumes is by construction, not by audit. Two frozen-gain 1.2B arms (Dense seed 72, Mixed seed 70) are repaired runs, restarted from an earlier checkpoint according to their run names.

\textbf{Precision recipes.} \emph{Canonical} (pure bf16): every parameter, including the residual-stream RMSNorm and QK-normalisation gains, is stored and updated in bf16 with bf16 optimiser state and no fp32 master copy; every normalisation gain read out stayed exactly at 1.0 (Appendix B; the QK gains of all 22 twin checkpoints and all 113 gains of one; the control arms' gains were not read out). \emph{Learnable QK gains}: the 56 \path{q_norm}/\path{k_norm} gain tensors are kept as fp32 master weights with fp32 AdamW moments and cast to bf16 inside the forward pass (training flag \path{qk_gain_fp32}); every other parameter keeps the canonical precision (checkpoints hold 56 fp32 and 255 bf16 tensors; the residual RMSNorm gains remain bf16 and frozen at 1.0). This recipe was trained at 365M for seeds 42--44 and at 1.2B for seeds 70--72 (run identifiers \path{FINAL_20260924}, \path{FINAL3_20260927} and \path{FINAL5_20261014} in the released archive). \emph{Standard recipe}: fp32 master weights and fp32 AdamW state for every parameter, bf16 autocast compute, round-to-nearest, no stochastic rounding (all 311 tensors fp32 in the checkpoint). It was trained at 365M for seeds 41--43 (run identifier \path{FINAL6_20261020/21}). Two further single seed-42 parameterisations: \emph{all-norm} keeps all 113 normalisation-gain tensors (56 Q/K, 56 residual RMSNorm, final norm) as fp32 master weights (113 fp32 and 198 bf16 tensors); \emph{scalar} leaves every gain vector canonical (bf16, 1.0, frozen) and adds one learnable fp32 scalar per Q/K norm module (56 scalars, initialised 1.0, same learning rate and weight decay) that multiplies the normalised queries, respectively keys. On checkpoints with learned gains the temperature divides the learned fp32 \path{q_norm} gain by $\tau$ from an unmodified fp32 copy; for the scalar pair it scales the queries by 1/$\tau$ (an exact fp32 attention scale), because dividing the bf16-cast scalar realises $\tau$ only to bf16 precision; the two paths agree exactly on the 365M checkpoints and to within 0.03\% in perplexity on the 1.2B Mixed seed-72 checkpoint.

\textbf{Mixed-arm masks.} Block size 128; a pool of 10,000 pre-generated causal block masks (\path{sparse_mask_pool_size=10000}); each pool mask draws its budget \emph{k} uniformly from \{63, 47, 31, 21, 11\} of 64 blocks at the 8,192-token training length (budgets 98\%, 73\%, 48\%, 33\%, 17\%; mean $\approx$ 54\%). At every training step each attention layer selects one pool entry uniformly at random, independently of the other layers, and applies it to every sequence in the batch. A query block's own (diagonal) block is always retained through the local causal path; the remaining retained blocks are sampled uniformly from the non-diagonal blocks and blocks in the future of the query block are removed by causality, so early query blocks have fewer than \emph{k} retained predecessors and block 0 is retained only when sampled. The Dense arm uses full causal attention. The mask is the only difference between arms.

\textbf{Control arms (365M, canonical recipe).} The Static and Resampled controls fix the budget at the mean Mixed budget (checkpoint tag \texttt{nrb35}, about 35 of 64 blocks): Static uses one fixed mask throughout training and Resampled fresh masks of the same budget; seeds 42--47. The attention-dropout control uses \emph{p} = 0.28 (seeds 42--46). The scalar RoPE-distance control (checkpoint tag \path{ropectl_a1p0}) is a Dense model trained with a one-parameter change to RoPE, designed to test whether a mean-distance account explains Mixed's behaviour; its exact parameterisation is not recorded in the result files (seeds 42--46). The ALiBi control (\path{ALIBI_20260921}; seeds 42--46) replaces RoPE by head-specific linear distance biases with slopes \(2^{-8h/n_q}\) for query head \emph{h} of \(n_q = 6\) \citep{press2022}, keeps QK-normalisation and every other setting of the 365M Dense recipe, and uses the Dense twin's data seed and order; the bias is added to the scaled content logit, so the temperature rescales the content term only. The learnable-gain ALiBi arms (\path{FINAL7_20261023}; seeds 42, 43) add \path{qk_gain_fp32} to that recipe and are paired with the \path{FINAL_20260924} learnable-gain Dense of the same seed. The two regime pairs (\path{REGIME_20260922}; seed 42, Dense and Mixed both retrained) change one setting of the 365M recipe: the no-QK-norm pair removes the \path{q_norm}/\path{k_norm} modules; the RoPE-10$^{4}$ pair sets the rotary base to 10$^{4}$. The Dense-16k control uses the learnable-gain Dense recipe at sequence length 16,384 for 22,000 steps at a global batch of 20 $\times$ 16,384 tokens, i.e.~the same 7.21B training tokens as the 8k runs. Its loader concatenates 8k sequences 2\emph{i} and 2\emph{i}+1 of the seed-42 order into 16k sequence \emph{i}, so by construction the control sees the tokens of its 8k twin, in half as many optimiser steps with twice as many tokens per step.

\textbf{Data and licences.} All models are pretrained on a local export of FineWeb-Edu \texttt{sample-100BT} \citep{penedo2024}: 3,027,764 training sequences of 8,192 tokens (24.80B tokens; a deterministic 95/5 split leaves 159,356 validation sequences). At the recorded step counts the 365M, 1.2B and 3B runs consume 7.21B, 24.80B and 49.15B tokens, i.e.~0.29, 1.00 and 1.98 passes over the training split. Evaluation corpora: FineWeb-Edu validation windows from the same export; PG-19 (\citealp{rae2020}; Apache 2.0); arXiv from the Common Pile release \path{common-pile/arxiv_papers} (\citealp{kandpal2025}; openly licensed papers processed from their LaTeX sources). FineWeb-Edu is released under ODC-By 1.0 and the Qwen3 tokeniser under Apache 2.0. The evaluation windows were exported as fixed token arrays (96 windows of 8,192 and 24 of 30,720 tokens per corpus, two export copies); the export records neither document identifiers nor a pinned dataset revision, so the exact source documents are not recoverable from the window files. The training and evaluation text is predominantly English; the corpora are used unmodified for perplexity scoring only and nothing is redistributed.

\section{Evaluation protocol, the AdamW/bf16 bound and provenance}\label{evaluation-protocol-the-adamwbf16-bound-and-provenance}

\textbf{Windows.} The temperature protocol uses disjoint calibration and held-out sets of equal size: 8, 4, 2 and 2 non-overlapping windows per corpus at 8k, 16k, 24k and 30k (65,528, 65,532, 49,150 and 61,438 tokens per set). Every calibrated 24k and 30k result therefore uses two held-out windows per corpus, six per pair. The \emph{36-window set} adds ten further held-out windows per corpus at 24k and 30k, disjoint from every calibration window; it was evaluated in run \path{RESOLUTION_20260921} (file \path{RESOLUTION_20260921_driftfree_all_rows.tsv}, the re-run with hash-verified window manifests, which supersedes \path{RESOLUTION_20260921_all_rows.tsv}). The later runs' 30k position-resolved (diagnostics) windows, eight per corpus, include the two calibration and the two held-out windows, so their calibrated position profiles are partly in-sample and are reported as diagnostics; the canonical pairs' calibrated profiles (Tables C.3a--C.3b) use the two held-out windows. The default-temperature diagnostics of §4.1 (Table C.1a; Fig. 2a, solid) use 24, 12, 8 and 8 windows per corpus. Position-resolved cells score eight equal bins of a 30k window per corpus, seed and arm (bin edges 1, 3,841, 7,681, 11,521, 15,361, 19,200, 23,040, 26,880 and 30,719 tokens). Windows are contiguous slices of a concatenated token stream and can cross document boundaries. Perplexity is exp of the token-weighted mean NLL; pooled values pool the three corpora, corpus-mean $\Delta$ averages per-corpus log-perplexity differences.

\textbf{Calibration and statistics.} $\tau^{*}$ is the arg-min over the 13-point grid of the summed NLL over the calibration windows of the three corpora, per model and length, and is applied to the held-out windows. Twins sharing a seed are paired and the seed is the unit of analysis; cells report the mean paired $\Delta$, its SD across seeds and the number of seeds favouring the change; exact two-sided sign-flip p-values are in the supplementary tables (their 0.031 floor at six seeds restates a 6/6 count). On the 36-window set we report the mean paired difference with a 95\% t-half-width and the number of windows favouring the change; these characterise one pair and treat windows cut from one token stream as independent. Seed-level 90\% two-one-sided-test (TOST) intervals of the calibrated 30k residual (Appendix C.2) lie within the margin if and only if both one-sided tests reject at the 5\% level.

\textbf{The AdamW/bf16 bound.} The bf16 spacing is \(2^{-7} \approx 0.0078\) just above 1.0 and \(2^{-8}\) just below, so under round-to-nearest an update smaller than 0.002 (downward) or 0.004 (upward) leaves a value of exactly 1.0 unchanged. With $\beta_{1}$ = 0.9 and $\beta_{2}$ = 0.95 the bias-corrected moment ratio of Adam \citep{kingma2015} is bounded for every gradient sequence by (1 $-\beta_{1}$)/$\surd$((1 $-\beta_{2}$)(1 $-\beta_{1}^{2}$/$\beta_{2}$)) $\approx$ 1.165 (Cauchy--Schwarz over the two exponential averages); the unit bound sometimes quoted for this case does not hold in general. A step is therefore at most 1.165 $\times$ 5$\times$\(10^{-4}\) $\approx$ 5.8$\times$\(10^{-4}\) at the peak learning rate, plus a decoupled weight-decay term of 5$\times$10$^{-6}$. Rounding the moments themselves to bf16 perturbs each by at most a relative 2$^{-8}$ per step, which is small against the 3.4-fold margin between this largest step and the 1.95$\times$10$^{-3}$ needed. For $\beta_{2}$ = 0.98, 0.99 and 0.999 the same bound is 1.70, 2.35 and 7.27, so a bf16 gain at 1.0 cannot move at peak learning rates below 1.15$\times$10$^{-3}$, 8.3$\times$\(10^{-4}\) and 2.7$\times$\(10^{-4}\) respectively (1.68$\times$10$^{-3}$ at $\beta_{2}$ = 0.95). Every possible update here is therefore below the 1.95$\times$10$^{-3}$ needed to move a bf16 value downward off 1.0, and a parameter initialised at exactly 1.0 and stored in bf16 is frozen for the whole run. The effect is identical in both arms and does not disturb the pairing. A CPU audit of the canonical 365M Dense seed-42 checkpoint confirms it: all 113 normalisation-gain tensors are stored in bf16 and every element is exactly 1.0. In every one of the 22 canonical checkpoints (365M seeds 42--47, 1.2B seeds 42 and 70--72, 3B seed 42) every stored QK-normalisation gain value is exactly 1.0.

\textbf{Provenance and reproducibility.} The canonical cells were produced on the first cluster (A100 and H100 GPUs) and keyed by protocol (tier, length, split, corpus, window set, temperature, evaluator); entries evaluated in more than one run agree to within 0.024\% relative perplexity, and no differing values were averaged. Models trained after the first allocation ended (the 1.2B learnable-gain pairs, the ablations, the Dense-16k control, the standard-recipe pairs and the learnable-gain ALiBi arms) were evaluated on two further clusters (H100 and A100 GPUs) with the same evaluator revision and the same window manifests (verified by hash). On each new cluster the evaluator re-scored frozen canonical checkpoints before the 1.2B and ALiBi runs; the all-113 and RoPE-10$^{4}$ ablations had no such check, because their reference checkpoints were not available on that cluster, and the standard-recipe baseline was compared with the canonical cells after the run. The 1.2B seed-70 reference cells reproduced to within 5$\times$\(10^{-4}\)\%. The per-corpus 30k calibration-window cells of the frozen references reproduced to within 0.02\% (seed 71 on H100: twelve cells, maximum 0.018\%; seed 72 on A100: twelve cells, 0.020\%), and the ALiBi seed-42 reference cells to within 0.020\%. The 24 frozen-baseline cells assembled before the standard-recipe run agree with the first-cluster values to within 0.04\% (maximum 0.038\% on one FineWeb-Edu cell; 17 of the 18 cells with a first-cluster reference within 0.03\%; the frozen Mixed seed-43 arm had not been evaluated under the dense protocol on the first cluster); the twelve seed-42 cells among them coincide exactly with the first cluster's drift-free re-scoring and so add no independent cross-hardware measurement. An A100 recheck of the seed-71 cells reproduced the first-cluster values exactly, consistent with a numerical-path effect (reduction order under bf16/TF32) rather than an evaluator defect, although the seed-72 check on A100 GPUs still differed by up to 0.020\%. The largest such discrepancy among the canonical reference cells, 0.04\% in perplexity, is 0.0004 nats against the 0.03 log-PPL resolution of §3.2. Both arms of each later Dense/Mixed pair were evaluated on the same hardware, so their paired differences carry no cross-hardware term; the learnable-gain ALiBi arms are paired with Dense rows from the earlier run, whose seed-42 cells the ALiBi evaluation reproduced exactly. The 1.2B learnable-gain pairs were trained across several interrupted launches, each resumed from the last checkpoint; the later ablation and control runs (the all-113 and scalar gain flags \path{norm_gain_fp32} and \path{qk_gain_mode}, the RoPE-10$^{4}$ pair with learnable gains and the Dense-16k loader) were run from an uncommitted working copy of the training revision, so no commit identifies their training code; the standard-recipe and learnable-gain ALiBi runs record their training commit. Exact package versions were not recorded alongside the result files; the released evaluation code names the packages it imports.

\section{Result tables}\label{result-tables}

Tables are grouped by label: C.1 the canonical frozen-gain pairs at the default and calibrated temperatures; C.2 the 36-window set and equivalence intervals; C.3 position-resolved perplexity; C.4 the learnable-gain pairs; C.5 the standard-recipe pairs; C.6 ALiBi against RoPE with learnable gains (the frozen-gain ALiBi means are in C.7); C.7 the in-house control arms; C.8 regime pairs, gain-parameterisation ablations and the Dense-16k control; C.9 resolution of the calibrated comparison and the positive control; C.10 data-free temperature rules; C.11 the learned normalisation gains. The tables are full-width floats and appear in label order.

\begin{table*}[tp]
\centering
\begingroup\footnotesize\setlength{\tabcolsep}{4pt}
\sbox{\tblbox}{\begin{tabular}{@{}llllrrrrll@{}}
\toprule
Tier & Ctx & Corpus & n & Dense & Mixed & $\Delta$ log-PPL & SD & Mixed better & p\tabularnewline
\midrule
365M & 8k & arXiv & 6 & 21.86 & 22.64 & +0.035 & 0.019 & 0/6 & 0.031\tabularnewline
365M & 8k & FineWeb & 6 & 17.87 & 18.19 & +0.017 & 0.009 & 0/6 & 0.031\tabularnewline
365M & 8k & PG-19 & 6 & 42.46 & 43.51 & +0.024 & 0.018 & 0/6 & 0.031\tabularnewline
365M & 16k & arXiv & 6 & 23.99 & 25.04 & +0.043 & 0.019 & 0/6 & 0.031\tabularnewline
365M & 16k & FineWeb & 6 & 19.99 & 19.97 & $-$0.001 & 0.011 & 4/6 & 0.781\tabularnewline
365M & 16k & PG-19 & 6 & 46.36 & 47.41 & +0.022 & 0.020 & 0/6 & 0.031\tabularnewline
365M & 24k & arXiv & 6 & 29.73 & 30.83 & +0.036 & 0.038 & 1/6 & 0.062\tabularnewline
365M & 24k & FineWeb & 6 & 26.52 & 24.13 & $-$0.094 & 0.023 & 6/6 & 0.031\tabularnewline
365M & 24k & PG-19 & 6 & 56.48 & 55.04 & $-$0.026 & 0.021 & 5/6 & 0.062\tabularnewline
365M & 30k & arXiv & 6 & 38.75 & 39.16 & +0.010 & 0.046 & 3/6 & 0.625\tabularnewline
365M & 30k & FineWeb & 6 & 36.70 & 31.72 & $-$0.146 & 0.032 & 6/6 & 0.031\tabularnewline
365M & 30k & PG-19 & 6 & 68.45 & 64.05 & $-$0.067 & 0.022 & 6/6 & 0.031\tabularnewline
1.2B & 8k & arXiv & 3 & 14.55 & 14.79 & +0.016 & 0.029 & 1/3 & --\tabularnewline
1.2B & 8k & FineWeb & 3 & 12.00 & 12.22 & +0.019 & 0.007 & 0/3 & --\tabularnewline
1.2B & 8k & PG-19 & 3 & 28.09 & 28.65 & +0.020 & 0.007 & 0/3 & --\tabularnewline
1.2B & 16k & arXiv & 3 & 15.92 & 16.12 & +0.013 & 0.049 & 1/3 & --\tabularnewline
1.2B & 16k & FineWeb & 3 & 13.26 & 13.12 & $-$0.010 & 0.011 & 2/3 & --\tabularnewline
1.2B & 16k & PG-19 & 3 & 30.60 & 30.76 & +0.005 & 0.010 & 1/3 & --\tabularnewline
1.2B & 24k & arXiv & 3 & 21.10 & 20.52 & $-$0.028 & 0.079 & 2/3 & --\tabularnewline
1.2B & 24k & FineWeb & 3 & 19.57 & 16.69 & $-$0.159 & 0.026 & 3/3 & --\tabularnewline
1.2B & 24k & PG-19 & 3 & 40.64 & 37.71 & $-$0.075 & 0.025 & 3/3 & --\tabularnewline
1.2B & 30k & arXiv & 3 & 32.71 & 28.34 & $-$0.144 & 0.070 & 3/3 & --\tabularnewline
1.2B & 30k & FineWeb & 3 & 32.97 & 25.10 & $-$0.273 & 0.043 & 3/3 & --\tabularnewline
1.2B & 30k & PG-19 & 3 & 58.22 & 48.86 & $-$0.175 & 0.061 & 3/3 & --\tabularnewline
3B & 8k & arXiv & 1 & 12.33 & 12.40 & +0.005 & -- & -- & --\tabularnewline
3B & 8k & FineWeb & 1 & 9.95 & 10.02 & +0.007 & -- & -- & --\tabularnewline
3B & 8k & PG-19 & 1 & 23.94 & 23.54 & $-$0.017 & -- & -- & --\tabularnewline
3B & 16k & arXiv & 1 & 13.39 & 12.95 & $-$0.033 & -- & -- & --\tabularnewline
3B & 16k & FineWeb & 1 & 10.78 & 10.52 & $-$0.024 & -- & -- & --\tabularnewline
3B & 16k & PG-19 & 1 & 25.93 & 24.65 & $-$0.051 & -- & -- & --\tabularnewline
3B & 24k & arXiv & 1 & 18.53 & 15.85 & $-$0.156 & -- & -- & --\tabularnewline
3B & 24k & FineWeb & 1 & 16.09 & 13.12 & $-$0.204 & -- & -- & --\tabularnewline
3B & 24k & PG-19 & 1 & 38.15 & 30.25 & $-$0.232 & -- & -- & --\tabularnewline
3B & 30k & arXiv & 1 & 30.06 & 22.31 & $-$0.298 & -- & -- & --\tabularnewline
\bottomrule
\end{tabular}}\ifdim\wd\tblbox>\linewidth\resizebox{\linewidth}{!}{\usebox{\tblbox}}\else\usebox{\tblbox}\fi
\endgroup
\caption*{\textbf{Table C.1a:} True dense attention at the default temperature ($\tau$~=~1): perplexity and paired $\Delta$. 365M: seeds 42--47; 1.2B: seeds 42, 71, 72 (Mixed-70 was not evaluated on these diagnostics windows; its true-dense held-out values are in the released tables); 3B: seed 42.}
\end{table*}
\begin{table*}[tp]
\centering
\begingroup\footnotesize\setlength{\tabcolsep}{4pt}
\sbox{\tblbox}{\begin{tabular}{@{}llllrrrrll@{}}
\toprule
Tier & Ctx & Corpus & n & Dense & Mixed & $\Delta$ log-PPL & SD & Mixed better & p\tabularnewline
\midrule
3B & 30k & FineWeb & 1 & 27.68 & 20.06 & $-$0.322 & -- & -- & --\tabularnewline
3B & 30k & PG-19 & 1 & 60.70 & 40.32 & $-$0.409 & -- & -- & --\tabularnewline
\bottomrule
\end{tabular}}\ifdim\wd\tblbox>\linewidth\resizebox{\linewidth}{!}{\usebox{\tblbox}}\else\usebox{\tblbox}\fi
\endgroup
\caption*{\textbf{Table C.1a:} (continued)}
\end{table*}

\begin{table*}[tp]
\centering
\begingroup\footnotesize\setlength{\tabcolsep}{4pt}
\sbox{\tblbox}{\begin{tabular}{@{}lllrrll@{}}
\toprule
Tier & Seed & Ctx & $\Delta$ default & $\Delta$ calibrated & 36-window $\Delta\pm$ 95\% & Mixed better (of 36)\tabularnewline
\midrule
365M & 42 & 8k & +0.008 & +0.000 & &\tabularnewline
365M & 42 & 30k & $-$0.063 & $-$0.016 & $-$0.014 $\pm$ 0.004 & 31/36\tabularnewline
1.2B & 42 & 8k & +0.018 & +0.015 & &\tabularnewline
1.2B & 42 & 30k & $-$0.147 & $-$0.002 & +0.001 $\pm$ 0.005 & 18/36\tabularnewline
1.2B & 70 & 8k & +0.011 & +0.005 & &\tabularnewline
1.2B & 70 & 30k & $-$0.378 & $-$0.004 & $-$0.006 $\pm$ 0.005 & 25/36\tabularnewline
1.2B & 71 & 8k & +0.005 & +0.003 & &\tabularnewline
1.2B & 71 & 30k & $-$0.239 & +0.001 & +0.006 $\pm$ 0.005 & 15/36\tabularnewline
1.2B & 72 & 8k & +0.029 & +0.025 & &\tabularnewline
1.2B & 72 & 30k & $-$0.171 & +0.025 & +0.021 $\pm$ 0.008 & 9/36\tabularnewline
3B & 42 & 8k & $-$0.011 & $-$0.014 & &\tabularnewline
3B & 42 & 30k & $-$0.368 & +0.003 & +0.010 $\pm$ 0.014 & 20/36\tabularnewline
\bottomrule
\end{tabular}}\ifdim\wd\tblbox>\linewidth\resizebox{\linewidth}{!}{\usebox{\tblbox}}\else\usebox{\tblbox}\fi
\endgroup
\caption*{\textbf{Table C.1b:} Held-out temperature calibration under the dense operator at 8k and 30k, the six calibrated pairs (365M seed 42, the four 1.2B pairs, 3B). $\Delta$: log PPL(Mixed) $-$ log PPL(Dense), three-corpus mean on disjoint held-out windows.}
\end{table*}

\begin{table*}[tp]
\centering
\begingroup\scriptsize\setlength{\tabcolsep}{3pt}
\sbox{\tblbox}{\begin{tabular}{@{}lllllll@{}}
\toprule
Length & Seed & $\Delta$ Dense 0.85 / Mixed 0.80 & $\Delta$ common 0.85 & $\Delta$ common 0.80 & Dense PPL 0.85 & Mixed PPL 0.80\tabularnewline
\midrule
16k & 42 & $-$0.017 & $-$0.038 & $-$0.072 & 25.88 & 25.45\tabularnewline
16k & 43 & +0.014 & $-$0.012 & $-$0.037 & 25.55 & 25.90\tabularnewline
16k & 44 & +0.018 & $-$0.008 & $-$0.027 & 25.41 & 25.87\tabularnewline
16k & 45 & +0.007 & $-$0.018 & $-$0.052 & 25.34 & 25.50\tabularnewline
16k & 46 & $-$0.011 & $-$0.035 & $-$0.079 & 25.81 & 25.54\tabularnewline
16k & 47 & +0.005 & $-$0.013 & $-$0.045 & 25.52 & 25.65\tabularnewline
16k & mean (Mixed better) & +0.003 (2/6) & $-$0.021 (6/6) & $-$0.052 (6/6) & 25.58 & 25.65\tabularnewline
24k & 42 & $-$0.015 & $-$0.026 & $-$0.059 & 32.30 & 31.83\tabularnewline
24k & 43 & +0.020 & +0.005 & $-$0.022 & 31.79 & 32.43\tabularnewline
24k & 44 & +0.014 & $-$0.002 & $-$0.021 & 31.88 & 32.34\tabularnewline
24k & 45 & +0.021 & +0.007 & $-$0.025 & 31.57 & 32.24\tabularnewline
24k & 46 & $-$0.002 & $-$0.018 & $-$0.061 & 32.00 & 31.93\tabularnewline
24k & 47 & +0.010 & +0.001 & $-$0.029 & 31.81 & 32.13\tabularnewline
24k & mean (Mixed better) & +0.008 (2/6) & $-$0.005 (3/6) & $-$0.036 (6/6) & 31.89 & 32.15\tabularnewline
30k & 42 & $-$0.016 & $-$0.016 & $-$0.048 & 35.31 & 34.75\tabularnewline
30k & 43 & +0.019 & +0.017 & $-$0.011 & 34.87 & 35.55\tabularnewline
30k & 44 & +0.013 & +0.007 & $-$0.013 & 34.99 & 35.44\tabularnewline
30k & 45 & +0.019 & +0.016 & $-$0.016 & 34.66 & 35.33\tabularnewline
30k & 46 & $-$0.004 & $-$0.010 & $-$0.049 & 35.01 & 34.87\tabularnewline
30k & 47 & +0.012 & +0.012 & $-$0.016 & 34.81 & 35.23\tabularnewline
30k & mean (Mixed better) & +0.007 (2/6) & +0.004 (2/6) & $-$0.025 (6/6) & 34.94 & 35.19\tabularnewline
\bottomrule
\end{tabular}}\ifdim\wd\tblbox>\linewidth\resizebox{\linewidth}{!}{\usebox{\tblbox}}\else\usebox{\tblbox}\fi
\endgroup
\caption*{\textbf{Table C.1c:} Six-seed 365M held-out perplexity at the selected 30k temperatures and at common temperatures (frozen-gain checkpoints; held-out windows of the low-temperature check, two per corpus at each of 16k, 24k and 30k (at 16k not the four-window held-out set of Appendix B); $\tau\in$ \{0.75, 0.80, 0.85\}; dense operator). $\Delta$~=~corpus-mean log PPL(Mixed) $-$ log PPL(Dense); PPL columns are pooled over the three corpora.}
\end{table*}

\begin{table*}[tp]
\centering
\begingroup\footnotesize\setlength{\tabcolsep}{4pt}
\sbox{\tblbox}{\begin{tabular}{@{}llllll@{}}
\toprule
Tier & Seed & Context & $\Delta$ default ($\tau$~=~1) & 95\% CI & $\tau^{*}$ Dense / Mixed\tabularnewline
\midrule
1.2B & 42 & 24k & $-$0.077 & {[}$-$0.099, $-$0.054{]} & 0.90 / 0.85\tabularnewline
1.2B & 42 & 30k & $-$0.163 & {[}$-$0.192, $-$0.134{]} & 0.85 / 0.80\tabularnewline
1.2B & 70 & 24k & $-$0.198 & {[}$-$0.225, $-$0.171{]} & 0.85 / 0.85\tabularnewline
1.2B & 70 & 30k & $-$0.394 & {[}$-$0.432, $-$0.356{]} & 0.85 / 0.80\tabularnewline
1.2B & 71 & 24k & $-$0.113 & {[}$-$0.123, $-$0.102{]} & 0.85 / 0.85\tabularnewline
1.2B & 71 & 30k & $-$0.237 & {[}$-$0.252, $-$0.222{]} & 0.85 / 0.80\tabularnewline
1.2B & 72 & 24k & $-$0.056 & {[}$-$0.086, $-$0.026{]} & 0.90 / 0.85\tabularnewline
1.2B & 72 & 30k & $-$0.177 & {[}$-$0.219, $-$0.136{]} & 0.85 / 0.80\tabularnewline
365M & 42 & 24k & $-$0.036 & {[}$-$0.054, $-$0.018{]} & 0.85 / 0.85\tabularnewline
365M & 42 & 30k & $-$0.073 & {[}$-$0.099, $-$0.047{]} & 0.85 / 0.80\tabularnewline
3B & 42 & 24k & $-$0.213 & {[}$-$0.233, $-$0.193{]} & 0.85 / 0.85\tabularnewline
3B & 42 & 30k & $-$0.375 & {[}$-$0.407, $-$0.342{]} & 0.85 / 0.80\tabularnewline
\bottomrule
\end{tabular}}\ifdim\wd\tblbox>\linewidth\resizebox{\linewidth}{!}{\usebox{\tblbox}}\else\usebox{\tblbox}\fi
\endgroup
\caption*{\textbf{Table C.2a:} (columns 1/2) Calibrated residuals on 12 held-out windows per corpus (file \path{RESOLUTION_20260921_driftfree_all_rows.tsv}; Appendix B).}
\end{table*}
\begin{table*}[tp]
\centering
\begingroup\footnotesize\setlength{\tabcolsep}{4pt}
\sbox{\tblbox}{\begin{tabular}{@{}llllll@{}}
\toprule
Tier & Seed & $\Delta$ calibrated & 95\% CI & Mixed better & $\Delta$ at common Dense $\tau^{*}$\tabularnewline
\midrule
1.2B & 42 & $-$0.012 & {[}$-$0.015, $-$0.009{]} & 35/36 & +0.005\tabularnewline
1.2B & 42 & +0.001 & {[}$-$0.005, +0.006{]} & 18/36 & +0.003\tabularnewline
1.2B & 70 & $-$0.015 & {[}$-$0.019, $-$0.011{]} & 34/36 & $-$0.015\tabularnewline
1.2B & 70 & $-$0.006 & {[}$-$0.011, $-$0.001{]} & 25/36 & $-$0.005\tabularnewline
1.2B & 71 & $-$0.014 & {[}$-$0.018, $-$0.010{]} & 33/36 & $-$0.014\tabularnewline
1.2B & 71 & +0.006 & {[}+0.001, +0.012{]} & 15/36 & +0.007\tabularnewline
1.2B & 72 & +0.011 & {[}+0.004, +0.018{]} & 14/36 & +0.029\tabularnewline
1.2B & 72 & +0.021 & {[}+0.013, +0.030{]} & 9/36 & +0.024\tabularnewline
365M & 42 & $-$0.023 & {[}$-$0.028, $-$0.018{]} & 33/36 & $-$0.023\tabularnewline
365M & 42 & $-$0.014 & {[}$-$0.019, $-$0.010{]} & 31/36 & $-$0.013\tabularnewline
3B & 42 & $-$0.022 & {[}$-$0.028, $-$0.015{]} & 31/36 & $-$0.022\tabularnewline
3B & 42 & +0.010 & {[}$-$0.004, +0.023{]} & 20/36 & +0.019\tabularnewline
\bottomrule
\end{tabular}}\ifdim\wd\tblbox>\linewidth\resizebox{\linewidth}{!}{\usebox{\tblbox}}\else\usebox{\tblbox}\fi
\endgroup
\caption*{\textbf{Table C.2a:} (columns 2/2) Calibrated residuals on 12 held-out windows per corpus (file \path{RESOLUTION_20260921_driftfree_all_rows.tsv}; Appendix B).}
\end{table*}

\textbf{Equivalence intervals for the calibrated 30k residual.} From the per-seed values of Table C.1c (365M, six pairs at the transferred temperatures Dense 0.85 / Mixed 0.80: $-$0.016, +0.019, +0.013, +0.019, $-$0.004, +0.012) the 90\% TOST interval is {[}$-$0.004, +0.019{]}; from the calibrated 30k column of Table C.1b (1.2B, four pairs: $-$0.002, $-$0.004, +0.001, +0.025) it is {[}$-$0.011, +0.021{]}. Both lie within the grid-step margin of $\pm$0.03 log-PPL (§3.2); the 365M interval also lies within $\pm$0.02, the 1.2B interval does not. Seed means of perplexity in the tables are arithmetic in Table C.1a and geometric in the mean rows of Tables C.1c and C.3a. The 3B pair admits no seed-level interval.

\begin{table*}[tp]
\centering
\begingroup\footnotesize\setlength{\tabcolsep}{4pt}
\sbox{\tblbox}{\begin{tabular}{@{}llllllllll@{}}
\toprule
Tier & Arm & bin 0 & bin 1 & bin 2 & bin 3 & bin 4 & bin 5 & bin 6 & bin 7\tabularnewline
\midrule
365M & dense & 41.84 & 33.56 & 40.10 & 32.59 & 38.45 & 33.55 & 32.41 & 31.51\tabularnewline
365M & mixed & 40.76 & 32.05 & 39.04 & 32.72 & 38.21 & 33.12 & 32.06 & 31.44\tabularnewline
1.2B & dense & 26.49 & 21.70 & 26.34 & 22.39 & 26.56 & 22.56 & 22.39 & 22.27\tabularnewline
1.2B & mixed & 26.59 & 21.40 & 26.31 & 22.46 & 26.68 & 22.88 & 22.77 & 22.60\tabularnewline
3B & dense & 22.22 & 17.97 & 22.26 & 19.12 & 22.85 & 18.96 & 18.99 & 19.86\tabularnewline
3B & mixed & 21.81 & 17.39 & 21.45 & 18.53 & 22.18 & 18.96 & 20.22 & 22.17\tabularnewline
\bottomrule
\end{tabular}}\ifdim\wd\tblbox>\linewidth\resizebox{\linewidth}{!}{\usebox{\tblbox}}\else\usebox{\tblbox}\fi
\endgroup
\caption*{\textbf{Table C.3a:} Absolute perplexity by position bin at the calibrated temperatures (geometric mean over corpora and windows; 365M: seed 42 at its own $\tau^{*}$; 1.2B: mean over the four pairs; 3B: seed 42). The alternation between odd and even bins reflects the content of the specific windows rather than position.}
\end{table*}

\begin{table*}[tp]
\centering
\begingroup\footnotesize\setlength{\tabcolsep}{4pt}
\sbox{\tblbox}{\begin{tabular}{@{}llllll@{}}
\toprule
Bin (30k window) & $\Delta$ Dense 0.85 / Mixed 0.80 & $\Delta$ common 0.85 & $\Delta$ common 0.80 & Dense PPL 0.85 & Mixed PPL 0.80\tabularnewline
\midrule
1 & +0.003 & $-$0.025 & $-$0.045 & 41.0 & 41.1\tabularnewline
2 & $-$0.011 & $-$0.041 & $-$0.078 & 32.9 & 32.5\tabularnewline
3 & $-$0.007 & $-$0.039 & $-$0.079 & 39.8 & 39.5\tabularnewline
4 & +0.020 & +0.002 & $-$0.025 & 32.4 & 33.1\tabularnewline
5 & +0.017 & +0.014 & $-$0.016 & 38.1 & 38.8\tabularnewline
6 & +0.016 & +0.029 & $-$0.000 & 33.1 & 33.7\tabularnewline
7 & +0.011 & +0.041 & +0.012 & 32.2 & 32.6\tabularnewline
8 & +0.009 & +0.052 & +0.028 & 31.4 & 31.7\tabularnewline
\bottomrule
\end{tabular}}\ifdim\wd\tblbox>\linewidth\resizebox{\linewidth}{!}{\usebox{\tblbox}}\else\usebox{\tblbox}\fi
\endgroup
\caption*{\textbf{Table C.3b:} Six-seed 365M position-resolved $\Delta$ log-PPL at 30k at the selected and at common temperatures (frozen-gain checkpoints; held-out windows, two per corpus; eight equal bins of $\approx$3.8k tokens; $\Delta$~=~corpus-mean log PPL(Mixed) $-$ log PPL(Dense), mean over six seeds; bins numbered 1--8 here, 0--7 elsewhere). The first column is the six-seed 365M calibrated per-bin series (seed 42 alone gives $-$0.026, $-$0.046, $-$0.027, +0.004, $-$0.006, $-$0.013, $-$0.011, $-$0.002; the supplementary tables).}
\end{table*}

\begin{table*}[tp]
\centering
\begingroup\footnotesize\setlength{\tabcolsep}{4pt}
\sbox{\tblbox}{\begin{tabular}{@{}llllllllllll@{}}
\toprule
Tier & Context & Corpus & n & bin 0 & bin 1 & bin 2 & bin 3 & bin 4 & bin 5 & bin 6 & bin 7\tabularnewline
\midrule
365M & 30k & arxiv & 6 & 0.030 & 0.036 & 0.060 & 0.035 & 0.013 & $-$0.019 & $-$0.052 & $-$0.021\tabularnewline
365M & 30k & fineweb & 6 & 0.014 & 0.019 & 0.029 & $-$0.039 & $-$0.155 & $-$0.303 & $-$0.362 & $-$0.371\tabularnewline
365M & 30k & pg19 & 6 & 0.016 & 0.025 & 0.046 & 0.003 & $-$0.064 & $-$0.143 & $-$0.197 & $-$0.219\tabularnewline
1.2B & 30k & arxiv & 3 & 0.017 & 0.023 & 0.025 & $-$0.004 & $-$0.074 & $-$0.157 & $-$0.381 & $-$0.604\tabularnewline
1.2B & 30k & fineweb & 3 & 0.008 & 0.019 & 0.024 & $-$0.057 & $-$0.227 & $-$0.464 & $-$0.641 & $-$0.846\tabularnewline
1.2B & 30k & pg19 & 3 & 0.014 & 0.023 & 0.031 & $-$0.027 & $-$0.142 & $-$0.263 & $-$0.429 & $-$0.610\tabularnewline
3B & 30k & arxiv & 1 & $-$0.002 & 0.009 & $-$0.002 & $-$0.104 & $-$0.321 & $-$0.477 & $-$0.659 & $-$0.829\tabularnewline
3B & 30k & fineweb & 1 & 0.005 & 0.013 & 0.008 & $-$0.076 & $-$0.281 & $-$0.553 & $-$0.816 & $-$0.873\tabularnewline
3B & 30k & pg19 & 1 & $-$0.020 & $-$0.014 & $-$0.012 & $-$0.095 & $-$0.327 & $-$0.621 & $-$0.991 & $-$1.193\tabularnewline
\bottomrule
\end{tabular}}\ifdim\wd\tblbox>\linewidth\resizebox{\linewidth}{!}{\usebox{\tblbox}}\else\usebox{\tblbox}\fi
\endgroup
\caption*{\textbf{Table C.3c:} Position-resolved $\Delta$ log-PPL at the default temperature (eight bins per window; 30k rows).}
\end{table*}

\begin{table*}[tp]
\centering
\begingroup\scriptsize\setlength{\tabcolsep}{3pt}
\sbox{\tblbox}{\begin{tabular}{@{}lllllll@{}}
\toprule
Seed & Length & $\tau^{*}$ Dense / Mixed & Dense PPL $\tau$~=~1 $\rightarrow\tau$* & Mixed PPL $\tau$~=~1 $\rightarrow\tau$* & $\Delta$ default & $\Delta$ calibrated\tabularnewline
\midrule
42 & 8k & 1.00 / 0.95 & 31.32 $\rightarrow$ 31.32 & 32.33 $\rightarrow$ 32.35 & +0.032 & +0.032\tabularnewline
43 & 8k & 1.00 / 0.95 & 31.82 $\rightarrow$ 31.82 & 32.52 $\rightarrow$ 32.55 & +0.022 & +0.022\tabularnewline
44 & 8k & 1.00 / 1.00 & 31.72 $\rightarrow$ 31.72 & 32.32 $\rightarrow$ 32.32 & +0.019 & +0.019\tabularnewline
mean (Mixed better) & 8k & -- & 31.62 $\rightarrow$ 31.62 & 32.39 $\rightarrow$ 32.40 & +0.024 (0/3) & +0.025 (0/3)\tabularnewline
42 & 16k & 0.95 / 0.95 & 31.14 $\rightarrow$ 31.06 & 32.42 $\rightarrow$ 32.10 & +0.040 & +0.033\tabularnewline
43 & 16k & 0.95 / 0.90 & 31.80 $\rightarrow$ 31.61 & 32.73 $\rightarrow$ 32.34 & +0.029 & +0.023\tabularnewline
44 & 16k & 0.95 / 0.95 & 31.84 $\rightarrow$ 31.72 & 32.65 $\rightarrow$ 32.30 & +0.025 & +0.018\tabularnewline
mean (Mixed better) & 16k & -- & 31.59 $\rightarrow$ 31.46 & 32.60 $\rightarrow$ 32.25 & +0.031 (0/3) & +0.025 (0/3)\tabularnewline
42 & 24k & 0.95 / 0.90 & 29.77 $\rightarrow$ 29.32 & 31.22 $\rightarrow$ 30.23 & +0.047 & +0.031\tabularnewline
43 & 24k & 0.90 / 0.90 & 30.55 $\rightarrow$ 29.88 & 31.77 $\rightarrow$ 30.61 & +0.039 & +0.024\tabularnewline
44 & 24k & 0.90 / 0.90 & 30.66 $\rightarrow$ 30.12 & 31.71 $\rightarrow$ 30.62 & +0.034 & +0.017\tabularnewline
mean (Mixed better) & 24k & -- & 30.33 $\rightarrow$ 29.77 & 31.57 $\rightarrow$ 30.49 & +0.040 (0/3) & +0.024 (0/3)\tabularnewline
42 & 30k & 0.90 / 0.90 & 33.13 $\rightarrow$ 32.34 & 34.88 $\rightarrow$ 33.29 & +0.051 & +0.029\tabularnewline
43 & 30k & 0.90 / 0.85 & 34.47 $\rightarrow$ 32.97 & 35.76 $\rightarrow$ 33.75 & +0.037 & +0.024\tabularnewline
44 & 30k & 0.90 / 0.90 & 34.40 $\rightarrow$ 33.23 & 35.71 $\rightarrow$ 33.80 & +0.037 & +0.017\tabularnewline
mean (Mixed better) & 30k & -- & 34.00 $\rightarrow$ 32.84 & 35.45 $\rightarrow$ 33.61 & +0.042 (0/3) & +0.023 (0/3)\tabularnewline
\bottomrule
\end{tabular}}\ifdim\wd\tblbox>\linewidth\resizebox{\linewidth}{!}{\usebox{\tblbox}}\else\usebox{\tblbox}\fi
\endgroup
\caption*{\textbf{Table C.4a:} The learnable-gain pairs at all four lengths (protocol calibration and held-out windows; pooled held-out perplexity; $\Delta$~=~log PPL(Mixed) $-$ log PPL(Dense), three-corpus mean). Per seed, with the three-seed mean and the number of pairs favouring Mixed.}
\end{table*}

\begin{table*}[tp]
\centering
\begingroup\footnotesize\setlength{\tabcolsep}{4pt}
\sbox{\tblbox}{\begin{tabular}{@{}lllll@{}}
\toprule
Seed & Length & Temperatures (Dense / Mixed) & $\Delta$ mean $\pm$ 95\% & Mixed better\tabularnewline
\midrule
42 & 24k & default (1.00 / 1.00) & +0.047 $\pm$ 0.005 & 0/36\tabularnewline
42 & 24k & calibrated (0.95 / 0.90) & +0.032 $\pm$ 0.004 & 0/36\tabularnewline
43 & 24k & default (1.00 / 1.00) & +0.038 $\pm$ 0.006 & 0/36\tabularnewline
43 & 24k & calibrated (0.90 / 0.90) & +0.023 $\pm$ 0.006 & 0/36\tabularnewline
44 & 24k & default (1.00 / 1.00) & +0.039 $\pm$ 0.004 & 0/36\tabularnewline
44 & 24k & calibrated (0.90 / 0.90) & +0.019 $\pm$ 0.004 & 0/36\tabularnewline
seed-averaged & 24k & default (per seed) & +0.041 $\pm$ 0.005 & 0/36\tabularnewline
seed-averaged & 24k & calibrated (per seed) & +0.025 $\pm$ 0.004 & 0/36\tabularnewline
42 & 30k & default (1.00 / 1.00) & +0.056 $\pm$ 0.006 & 0/36\tabularnewline
42 & 30k & calibrated (0.90 / 0.90) & +0.034 $\pm$ 0.005 & 0/36\tabularnewline
43 & 30k & default (1.00 / 1.00) & +0.043 $\pm$ 0.007 & 0/36\tabularnewline
43 & 30k & calibrated (0.90 / 0.85) & +0.028 $\pm$ 0.005 & 0/36\tabularnewline
44 & 30k & default (1.00 / 1.00) & +0.045 $\pm$ 0.005 & 0/36\tabularnewline
44 & 30k & calibrated (0.90 / 0.90) & +0.024 $\pm$ 0.004 & 0/36\tabularnewline
seed-averaged & 30k & default (per seed) & +0.048 $\pm$ 0.006 & 0/36\tabularnewline
seed-averaged & 30k & calibrated (per seed) & +0.029 $\pm$ 0.004 & 0/36\tabularnewline
\bottomrule
\end{tabular}}\ifdim\wd\tblbox>\linewidth\resizebox{\linewidth}{!}{\usebox{\tblbox}}\else\usebox{\tblbox}\fi
\endgroup
\caption*{\textbf{Table C.4b:} The learnable-gain pairs on the 36 resolution windows (12 per corpus; the windows of Table C.2a), 24k and 30k. Mean paired $\Delta$ log-PPL (Mixed~$-$~Dense) $\pm$ 95\% t-half-width over windows, windows favouring Mixed, per seed and for the seed-averaged per-window difference (each arm at its own $\tau^{*}$ per seed).}
\end{table*}

\begin{table*}[tp]
\centering
\begingroup\footnotesize\setlength{\tabcolsep}{4pt}
\sbox{\tblbox}{\begin{tabular}{@{}lllll@{}}
\toprule
Seed & Arm & Length & Frozen gains: $\tau^{*}$, PPL $\tau$~=~1 $\rightarrow\tau$* (gain) & Learnable gains: $\tau^{*}$, PPL $\tau$~=~1 $\rightarrow\tau$* (gain)\tabularnewline
\midrule
42 & Dense & 8k & 1.00, 32.8 $\rightarrow$ 32.8 (0\%) & 1.00, 31.3 $\rightarrow$ 31.3 (0\%)\tabularnewline
42 & Dense & 30k & 0.85, 55.9 $\rightarrow$ 35.3 (37\%) & 0.90, 33.1 $\rightarrow$ 32.3 (2\%)\tabularnewline
42 & Mixed & 8k & 0.95, 33.1 $\rightarrow$ 32.9 (1\%) & 0.95, 32.3 $\rightarrow$ 32.3 (0\%)\tabularnewline
42 & Mixed & 30k & 0.80, 52.5 $\rightarrow$ 34.8 (34\%) & 0.90, 34.9 $\rightarrow$ 33.3 (5\%)\tabularnewline
43 & Dense & 8k & 1.00, 32.7 $\rightarrow$ 32.7 (0\%) & 1.00, 31.8 $\rightarrow$ 31.8 (0\%)\tabularnewline
43 & Dense & 30k & 0.85, 56.9 $\rightarrow$ 34.9 (39\%) & 0.90, 34.5 $\rightarrow$ 33.0 (4\%)\tabularnewline
43 & Mixed & 8k & -- & 0.95, 32.5 $\rightarrow$ 32.5 (0\%)\tabularnewline
43 & Mixed & 30k & -- & 0.85, 35.8 $\rightarrow$ 33.8 (6\%)\tabularnewline
44 & Dense & 8k & 1.00, 32.8 $\rightarrow$ 32.8 (0\%) & 1.00, 31.7 $\rightarrow$ 31.7 (0\%)\tabularnewline
44 & Dense & 30k & 0.85, 56.1 $\rightarrow$ 35.0 (38\%) & 0.90, 34.4 $\rightarrow$ 33.2 (3\%)\tabularnewline
44 & Mixed & 8k & -- & 1.00, 32.3 $\rightarrow$ 32.3 (0\%)\tabularnewline
44 & Mixed & 30k & -- & 0.90, 35.7 $\rightarrow$ 33.8 (5\%)\tabularnewline
\bottomrule
\end{tabular}}\ifdim\wd\tblbox>\linewidth\resizebox{\linewidth}{!}{\usebox{\tblbox}}\else\usebox{\tblbox}\fi
\endgroup
\caption*{\textbf{Table C.4c:} (columns 1/2) Frozen against learnable gains for the same seeds at 8k and 30k (pooled held-out perplexity on the protocol windows). Frozen: the canonical checkpoints (scored under the temperature protocol; the Mixed seed-43 and seed-44 cells are absent from the canonical cell table and are left blank); learnable: the \path{FINAL_20260924} checkpoints. ``Gain'' is the perplexity reduction from $\tau$~=~1 to $\tau^{*}$.}
\end{table*}
\begin{table*}[tp]
\centering
\begingroup\footnotesize\setlength{\tabcolsep}{4pt}
\sbox{\tblbox}{\begin{tabular}{@{}llll@{}}
\toprule
Seed & Arm & PPL ratio learnable / frozen at $\tau$~=~1 & at $\tau^{*}$\tabularnewline
\midrule
42 & Dense & 0.954 & 0.954\tabularnewline
42 & Dense & 0.592 & 0.916\tabularnewline
42 & Mixed & 0.977 & 0.985\tabularnewline
42 & Mixed & 0.664 & 0.958\tabularnewline
43 & Dense & 0.974 & 0.974\tabularnewline
43 & Dense & 0.605 & 0.945\tabularnewline
43 & Mixed & -- & --\tabularnewline
43 & Mixed & -- & --\tabularnewline
44 & Dense & 0.966 & 0.966\tabularnewline
44 & Dense & 0.613 & 0.950\tabularnewline
44 & Mixed & -- & --\tabularnewline
44 & Mixed & -- & --\tabularnewline
\bottomrule
\end{tabular}}\ifdim\wd\tblbox>\linewidth\resizebox{\linewidth}{!}{\usebox{\tblbox}}\else\usebox{\tblbox}\fi
\endgroup
\caption*{\textbf{Table C.4c:} (columns 2/2) Frozen against learnable gains for the same seeds at 8k and 30k (pooled held-out perplexity on the protocol windows). Frozen: the canonical checkpoints (scored under the temperature protocol; the Mixed seed-43 and seed-44 cells are absent from the canonical cell table and are left blank); learnable: the \path{FINAL_20260924} checkpoints. ``Gain'' is the perplexity reduction from $\tau$~=~1 to $\tau^{*}$.}
\end{table*}

\begin{table*}[tp]
\centering
\begingroup\footnotesize\setlength{\tabcolsep}{4pt}
\sbox{\tblbox}{\begin{tabular}{@{}lllll@{}}
\toprule
Pair & Length & $\tau^{*}$ Dense / Mixed & Dense PPL $\tau$~=~1 $\rightarrow\tau$* (gain) & Mixed PPL $\tau$~=~1 $\rightarrow\tau$* (gain)\tabularnewline
\midrule
seed 70, frozen gains & 30k & 0.85 / 0.80 & 56.11 $\rightarrow$ 23.81 (57.6\%) & 38.45 $\rightarrow$ 23.72 (38.3\%)\tabularnewline
seed 70, learnable gains & 8k & 1.00 / 0.95 & 21.70 $\rightarrow$ 21.70 (0.0\%) & 21.83 $\rightarrow$ 21.85 ($-$0.1\%)\tabularnewline
seed 70, learnable gains & 16k & 0.95 / 0.95 & 21.79 $\rightarrow$ 21.51 (1.3\%) & 21.75 $\rightarrow$ 21.60 (0.7\%)\tabularnewline
seed 70, learnable gains & 24k & 0.90 / 0.90 & 21.72 $\rightarrow$ 20.28 (6.7\%) & 20.74 $\rightarrow$ 20.34 (1.9\%)\tabularnewline
seed 70, learnable gains & 30k & 0.90 / 0.90 & 25.23 $\rightarrow$ 22.41 (11.2\%) & 23.04 $\rightarrow$ 22.39 (2.8\%)\tabularnewline
seed 71, frozen gains & 30k & 0.85 / 0.80 & 49.91 $\rightarrow$ 23.93 (52.1\%) & 39.32 $\rightarrow$ 23.96 (39.1\%)\tabularnewline
seed 71, learnable gains & 8k & 1.00 / 1.00 & 21.67 $\rightarrow$ 21.67 (0.0\%) & 21.75 $\rightarrow$ 21.75 (0.0\%)\tabularnewline
seed 71, learnable gains & 16k & 0.95 / 0.95 & 21.77 $\rightarrow$ 21.60 (0.8\%) & 21.60 $\rightarrow$ 21.49 (0.5\%)\tabularnewline
seed 71, learnable gains & 24k & 0.90 / 0.90 & 21.24 $\rightarrow$ 20.48 (3.6\%) & 20.43 $\rightarrow$ 20.20 (1.2\%)\tabularnewline
seed 71, learnable gains & 30k & 0.90 / 0.90 & 24.33 $\rightarrow$ 22.66 (6.9\%) & 22.72 $\rightarrow$ 22.30 (1.9\%)\tabularnewline
seed 72, frozen gains & 30k & 0.85 / 0.80 & 48.03 $\rightarrow$ 23.29 (51.5\%) & 40.46 $\rightarrow$ 23.88 (41.0\%)\tabularnewline
seed 72, learnable gains & 8k & 1.00 / 1.00 & 21.80 $\rightarrow$ 21.80 (0.0\%) & 22.01 $\rightarrow$ 22.01 (0.0\%)\tabularnewline
seed 72, learnable gains & 16k & 0.95 / 0.95 & 21.74 $\rightarrow$ 21.69 (0.2\%) & 21.88 $\rightarrow$ 21.77 (0.5\%)\tabularnewline
seed 72, learnable gains & 24k & 0.95 / 0.90 & 20.68 $\rightarrow$ 20.31 (1.8\%) & 20.65 $\rightarrow$ 20.31 (1.6\%)\tabularnewline
seed 72, learnable gains & 30k & 0.90 / 0.90 & 23.45 $\rightarrow$ 22.68 (3.3\%) & 23.01 $\rightarrow$ 22.46 (2.4\%)\tabularnewline
mean of 3 (learnable) & 8k &  & Dense gain 0.0\% & \tabularnewline
mean of 3 (learnable) & 16k &  & Dense gain 0.8\% & \tabularnewline
mean of 3 (learnable) & 24k &  & Dense gain 4.0\% & \tabularnewline
mean of 3 (learnable) & 30k &  & Dense gain 7.1\% & \tabularnewline
\bottomrule
\end{tabular}}\ifdim\wd\tblbox>\linewidth\resizebox{\linewidth}{!}{\usebox{\tblbox}}\else\usebox{\tblbox}\fi
\endgroup
\caption*{\textbf{Table C.4d:} (columns 1/2) 1.2B learnable-Q/K-gain pairs (seeds 70, 71, 72), dense operator, FINAL\_20260924 protocol. Frozen twins from the canonical table for comparison (30k only).}
\end{table*}
\begin{table*}[tp]
\centering
\begingroup\footnotesize\setlength{\tabcolsep}{4pt}
\sbox{\tblbox}{\begin{tabular}{@{}llll@{}}
\toprule
Pair & Length & $\Delta$ default & $\Delta$ calibrated\tabularnewline
\midrule
seed 70, frozen gains & 30k & $-$0.378 & $-$0.004\tabularnewline
seed 70, learnable gains & 8k & +0.006 & +0.007\tabularnewline
seed 70, learnable gains & 16k & $-$0.002 & +0.004\tabularnewline
seed 70, learnable gains & 24k & $-$0.046 & +0.003\tabularnewline
seed 70, learnable gains & 30k & $-$0.091 & $-$0.001\tabularnewline
seed 71, frozen gains & 30k & $-$0.239 & +0.001\tabularnewline
seed 71, learnable gains & 8k & +0.004 & +0.004\tabularnewline
seed 71, learnable gains & 16k & $-$0.008 & $-$0.005\tabularnewline
seed 71, learnable gains & 24k & $-$0.039 & $-$0.014\tabularnewline
seed 71, learnable gains & 30k & $-$0.068 & $-$0.016\tabularnewline
seed 72, frozen gains & 30k & $-$0.171 & +0.025\tabularnewline
seed 72, learnable gains & 8k & +0.010 & +0.010\tabularnewline
seed 72, learnable gains & 16k & +0.006 & +0.003\tabularnewline
seed 72, learnable gains & 24k & $-$0.001 & $-$0.000\tabularnewline
seed 72, learnable gains & 30k & $-$0.019 & $-$0.010\tabularnewline
mean of 3 (learnable) & 8k & +0.006 (SD 0.003; Mixed ahead 0/3) & +0.007 (SD 0.003; within $\pm$0.03 3/3)\tabularnewline
mean of 3 (learnable) & 16k & $-$0.001 (SD 0.007; Mixed ahead 2/3) & +0.001 (SD 0.005; within $\pm$0.03 3/3)\tabularnewline
mean of 3 (learnable) & 24k & $-$0.029 (SD 0.024; Mixed ahead 3/3) & $-$0.004 (SD 0.009; within $\pm$0.03 3/3)\tabularnewline
mean of 3 (learnable) & 30k & $-$0.059 (SD 0.037; Mixed ahead 3/3) & $-$0.009 (SD 0.008; within $\pm$0.03 3/3)\tabularnewline
\bottomrule
\end{tabular}}\ifdim\wd\tblbox>\linewidth\resizebox{\linewidth}{!}{\usebox{\tblbox}}\else\usebox{\tblbox}\fi
\endgroup
\caption*{\textbf{Table C.4d:} (columns 2/2) 1.2B learnable-Q/K-gain pairs (seeds 70, 71, 72), dense operator, FINAL\_20260924 protocol. Frozen twins from the canonical table for comparison (30k only).}
\end{table*}

\begin{table*}[tp]
\centering
\begingroup\footnotesize\setlength{\tabcolsep}{4pt}
\sbox{\tblbox}{\begin{tabular}{@{}lllll@{}}
\toprule
Pair & Length & Temperatures & $\Delta$ mean $\pm$ 95\% & Mixed better\tabularnewline
\midrule
seed 70, learnable gains & 24k & default (1.00 / 1.00) & $-$0.034 $\pm$ 0.006 & 36/36\tabularnewline
seed 70, learnable gains & 24k & calibrated (0.90 / 0.90) & +0.010 $\pm$ 0.004 & 7/36\tabularnewline
seed 70, learnable gains & 30k & default (1.00 / 1.00) & $-$0.077 $\pm$ 0.008 & 36/36\tabularnewline
seed 70, learnable gains & 30k & calibrated (0.90 / 0.90) & +0.005 $\pm$ 0.004 & 10/36\tabularnewline
seed 71, learnable gains & 24k & default (1.00 / 1.00) & $-$0.033 $\pm$ 0.011 & 35/36\tabularnewline
seed 71, learnable gains & 24k & calibrated (0.90 / 0.90) & $-$0.013 $\pm$ 0.007 & 26/36\tabularnewline
seed 71, learnable gains & 30k & default (1.00 / 1.00) & $-$0.058 $\pm$ 0.011 & 36/36\tabularnewline
seed 71, learnable gains & 30k & calibrated (0.90 / 0.90) & $-$0.015 $\pm$ 0.007 & 24/36\tabularnewline
seed 72, learnable gains & 24k & default (1.00 / 1.00) & +0.001 $\pm$ 0.007 & 15/36\tabularnewline
seed 72, learnable gains & 24k & calibrated (0.95 / 0.90) & +0.003 $\pm$ 0.005 & 15/36\tabularnewline
seed 72, learnable gains & 30k & default (1.00 / 1.00) & $-$0.011 $\pm$ 0.008 & 22/36\tabularnewline
seed 72, learnable gains & 30k & calibrated (0.90 / 0.90) & $-$0.004 $\pm$ 0.005 & 22/36\tabularnewline
\bottomrule
\end{tabular}}\ifdim\wd\tblbox>\linewidth\resizebox{\linewidth}{!}{\usebox{\tblbox}}\else\usebox{\tblbox}\fi
\endgroup
\caption*{\textbf{Table C.4e:} 1.2B learnable-gain pairs on the 36-window set (24k, 30k): Mixed~$-$~Dense, paired 95\% intervals, windows with Mixed ahead.}
\end{table*}

\begin{table*}[tp]
\centering
\begingroup\scriptsize\setlength{\tabcolsep}{3pt}
\sbox{\tblbox}{\begin{tabular}{@{}lllllll@{}}
\toprule
Bin (tokens) & Dense $\tau$~=~1 & Mixed $\tau$~=~1 & $\Delta\tau$ = 1 (seeds Mixed worse) & Dense $\tau^{*}$ & Mixed $\tau^{*}$ & $\Delta$ calibrated (seeds Mixed worse)\tabularnewline
\midrule
0 (0--3,840) & 27.3 & 28.1 & +0.026 (3/3) & 27.9 & 28.4 & +0.019 (3/3)\tabularnewline
1 (3,840--7,680) & 21.8 & 22.4 & +0.027 (3/3) & 22.3 & 22.7 & +0.019 (3/3)\tabularnewline
2 (7,680--11,520) & 25.2 & 25.9 & +0.030 (3/3) & 25.5 & 26.0 & +0.017 (3/3)\tabularnewline
3 (11,520--15,360) & 27.3 & 28.6 & +0.048 (3/3) & 26.6 & 27.4 & +0.030 (3/3)\tabularnewline
4 (15,360--19,200) & 28.0 & 29.7 & +0.057 (3/3) & 26.6 & 27.6 & +0.036 (3/3)\tabularnewline
5 (19,200--23,040) & 25.4 & 27.0 & +0.061 (3/3) & 23.7 & 24.5 & +0.034 (3/3)\tabularnewline
6 (23,040--26,880) & 28.4 & 30.4 & +0.070 (3/3) & 25.7 & 26.7 & +0.040 (3/3)\tabularnewline
7 (26,880--30,720) & 31.2 & 33.6 & +0.074 (3/3) & 27.5 & 28.7 & +0.043 (3/3)\tabularnewline
\bottomrule
\end{tabular}}\ifdim\wd\tblbox>\linewidth\resizebox{\linewidth}{!}{\usebox{\tblbox}}\else\usebox{\tblbox}\fi
\endgroup
\caption*{\textbf{Table C.4f:} Position-resolved perplexity at 30k for the learnable-gain pairs (eight equal bins of 3,840 tokens; geometric mean over corpora and the eight diagnostics windows per corpus; mean over the three seeds). Each arm at $\tau$~=~1 and at its own $\tau^{*}$ (Table C.4a); $\Delta$~=~log PPL(Mixed) $-$ log PPL(Dense) per bin, with the number of seeds in which Mixed is worse.}
\end{table*}

\begin{table*}[tp]
\centering
\begingroup\footnotesize\setlength{\tabcolsep}{4pt}
\sbox{\tblbox}{\begin{tabular}{@{}lllllllllll@{}}
\toprule
Pair & $\tau$ & bin 0 & bin 1 & bin 2 & bin 3 & bin 4 & bin 5 & bin 6 & bin 7 & last/first A / B\tabularnewline
\midrule
1.2B s70 learnable & $\tau$~=~1 & +0.006 & +0.014 & +0.014 & $-$0.009 & $-$0.076 & $-$0.138 & $-$0.192 & $-$0.270 & $\times$1.44 / $\times$1.09\tabularnewline
1.2B s70 learnable & $\tau^{*}$ & $-$0.002 & +0.003 & +0.004 & +0.013 & +0.006 & +0.001 & $-$0.002 & $-$0.006 & $\times$1.00 / $\times$0.99\tabularnewline
1.2B s71 learnable & $\tau$~=~1 & +0.007 & +0.007 & +0.001 & $-$0.023 & $-$0.049 & $-$0.090 & $-$0.141 & $-$0.183 & $\times$1.28 / $\times$1.06\tabularnewline
1.2B s71 learnable & $\tau^{*}$ & $-$0.001 & $-$0.003 & $-$0.008 & $-$0.010 & $-$0.012 & $-$0.026 & $-$0.023 & $-$0.031 & $\times$1.02 / $\times$0.99\tabularnewline
1.2B s72 learnable & $\tau$~=~1 & +0.010 & +0.020 & +0.015 & +0.016 & $-$0.008 & $-$0.001 & $-$0.012 & $-$0.080 & $\times$1.17 / $\times$1.07\tabularnewline
1.2B s72 learnable & $\tau^{*}$ & +0.003 & +0.011 & +0.002 & +0.005 & $-$0.010 & +0.006 & +0.006 & $-$0.012 & $\times$1.00 / $\times$0.98\tabularnewline
365M std s41 & $\tau$~=~1 & +0.025 & +0.026 & +0.021 & +0.000 & $-$0.003 & $-$0.010 & $-$0.043 & $-$0.061 & $\times$1.44 / $\times$1.32\tabularnewline
365M std s41 & $\tau^{*}$ & +0.023 & +0.026 & +0.010 & +0.005 & +0.004 & $-$0.009 & $-$0.028 & $-$0.038 & $\times$1.08 / $\times$1.01\tabularnewline
365M std s42 & $\tau$~=~1 & +0.023 & +0.022 & +0.017 & $-$0.001 & +0.013 & +0.026 & $-$0.000 & $-$0.030 & $\times$1.38 / $\times$1.31\tabularnewline
365M std s42 & $\tau^{*}$ & +0.020 & +0.020 & +0.007 & $-$0.001 & +0.010 & +0.008 & $-$0.013 & $-$0.023 & $\times$1.04 / $\times$1.00\tabularnewline
365M std s43 & $\tau$~=~1 & +0.020 & +0.017 & +0.022 & $-$0.026 & $-$0.081 & $-$0.106 & $-$0.167 & $-$0.220 & $\times$1.78 / $\times$1.40\tabularnewline
365M std s43 & $\tau^{*}$ & $-$0.001 & $-$0.006 & +0.008 & +0.022 & +0.014 & +0.010 & +0.003 & $-$0.009 & $\times$1.05 / $\times$1.04\tabularnewline
ALiBi s42 learnable & $\tau$~=~1 & +0.051 & +0.064 & +0.122 & +0.130 & +0.113 & +0.107 & +0.073 & +0.048 & $\times$1.09 / $\times$1.08\tabularnewline
ALiBi s42 learnable & $\tau^{*}$ & +0.033 & +0.044 & +0.095 & +0.149 & +0.152 & +0.162 & +0.153 & +0.141 & $\times$0.97 / $\times$1.08\tabularnewline
ALiBi s43 learnable & $\tau$~=~1 & +0.038 & +0.051 & +0.094 & +0.081 & +0.055 & +0.054 & $-$0.011 & $-$0.055 & $\times$1.17 / $\times$1.06\tabularnewline
ALiBi s43 learnable & $\tau^{*}$ & +0.018 & +0.032 & +0.084 & +0.111 & +0.117 & +0.135 & +0.105 & +0.089 & $\times$0.99 / $\times$1.06\tabularnewline
\bottomrule
\end{tabular}}\ifdim\wd\tblbox>\linewidth\resizebox{\linewidth}{!}{\usebox{\tblbox}}\else\usebox{\tblbox}\fi
\endgroup
\caption*{\textbf{Table C.4g:} Position-resolved $\Delta$ at 30k for the later pairs (eight bins, corpus means; $\Delta$~=~log PPL(B) $-$ log PPL(A), A = Dense, B = Mixed or ALiBi). Last column: last-bin / first-bin perplexity ratio of A and B at the row's temperatures.}
\end{table*}

\begin{table*}[tp]
\centering
\begingroup\footnotesize\setlength{\tabcolsep}{4pt}
\sbox{\tblbox}{\begin{tabular}{@{}lllll@{}}
\toprule
Seed & Length & $\tau^{*}$ Dense / Mixed & Dense PPL $\tau$~=~1 $\rightarrow\tau$* (gain) & Mixed PPL $\tau$~=~1 $\rightarrow\tau$* (gain)\tabularnewline
\midrule
41 & 8k & 1.00 / 0.95 & 29.18 $\rightarrow$ 29.18 (0.0\%) & 30.10 $\rightarrow$ 30.12 (0.0\%)\tabularnewline
41 & 16k & 0.95 / 0.90 & 30.25 $\rightarrow$ 29.81 (1.5\%) & 30.83 $\rightarrow$ 30.25 (1.9\%)\tabularnewline
41 & 24k & 0.90 / 0.90 & 29.81 $\rightarrow$ 28.32 (5.0\%) & 29.94 $\rightarrow$ 28.43 (5.1\%)\tabularnewline
41 & 30k & 0.90 / 0.85 & 34.56 $\rightarrow$ 31.57 (8.6\%) & 34.43 $\rightarrow$ 31.65 (8.1\%)\tabularnewline
42 & 8k & 1.00 / 0.95 & 29.54 $\rightarrow$ 29.54 (0.0\%) & 30.15 $\rightarrow$ 30.15 (0.0\%)\tabularnewline
42 & 16k & 0.95 / 0.90 & 30.54 $\rightarrow$ 30.14 (1.3\%) & 30.71 $\rightarrow$ 30.11 (2.0\%)\tabularnewline
42 & 24k & 0.90 / 0.90 & 29.80 $\rightarrow$ 28.43 (4.6\%) & 30.19 $\rightarrow$ 28.62 (5.2\%)\tabularnewline
42 & 30k & 0.90 / 0.85 & 34.55 $\rightarrow$ 31.73 (8.2\%) & 34.47 $\rightarrow$ 31.65 (8.2\%)\tabularnewline
43 & 8k & 1.00 / 0.95 & 29.72 $\rightarrow$ 29.72 (0.0\%) & 30.32 $\rightarrow$ 30.35 ($-$0.1\%)\tabularnewline
43 & 16k & 0.90 / 0.90 & 31.14 $\rightarrow$ 30.30 (2.7\%) & 31.27 $\rightarrow$ 30.64 (2.0\%)\tabularnewline
43 & 24k & 0.85 / 0.90 & 31.82 $\rightarrow$ 28.93 (9.1\%) & 30.76 $\rightarrow$ 28.99 (5.7\%)\tabularnewline
43 & 30k & 0.85 / 0.85 & 37.87 $\rightarrow$ 32.29 (14.7\%) & 35.32 $\rightarrow$ 32.33 (8.5\%)\tabularnewline
mean of 3 & 8k &  & Dense gain 0.0\% & \tabularnewline
mean of 3 & 16k &  & Dense gain 1.8\% & \tabularnewline
mean of 3 & 24k &  & Dense gain 6.2\% & \tabularnewline
mean of 3 & 30k &  & Dense gain 10.5\% & \tabularnewline
\bottomrule
\end{tabular}}\ifdim\wd\tblbox>\linewidth\resizebox{\linewidth}{!}{\usebox{\tblbox}}\else\usebox{\tblbox}\fi
\endgroup
\caption*{\textbf{Table C.5a:} (columns 1/2) 365M Dense/Mixed pairs under the standard mixed-precision recipe (fp32 master weights and fp32 AdamW state for every parameter; bf16 autocast compute), seeds 41--43, dense operator.}
\end{table*}
\begin{table*}[tp]
\centering
\begingroup\footnotesize\setlength{\tabcolsep}{4pt}
\sbox{\tblbox}{\begin{tabular}{@{}llll@{}}
\toprule
Seed & Length & $\Delta$ default & $\Delta$ calibrated\tabularnewline
\midrule
41 & 8k & +0.031 & +0.032\tabularnewline
41 & 16k & +0.019 & +0.015\tabularnewline
41 & 24k & +0.005 & +0.004\tabularnewline
41 & 30k & $-$0.004 & +0.003\tabularnewline
42 & 8k & +0.021 & +0.021\tabularnewline
42 & 16k & +0.006 & $-$0.001\tabularnewline
42 & 24k & +0.013 & +0.007\tabularnewline
42 & 30k & $-$0.002 & $-$0.003\tabularnewline
43 & 8k & +0.020 & +0.021\tabularnewline
43 & 16k & +0.004 & +0.011\tabularnewline
43 & 24k & $-$0.034 & +0.002\tabularnewline
43 & 30k & $-$0.070 & +0.001\tabularnewline
mean of 3 & 8k & +0.024 (SD 0.006; Mixed ahead 0/3) & +0.024 (SD 0.006; Mixed ahead 0/3)\tabularnewline
mean of 3 & 16k & +0.010 (SD 0.008; Mixed ahead 0/3) & +0.008 (SD 0.008; Mixed ahead 1/3)\tabularnewline
mean of 3 & 24k & $-$0.005 (SD 0.025; Mixed ahead 1/3) & +0.004 (SD 0.002; Mixed ahead 0/3)\tabularnewline
mean of 3 & 30k & $-$0.025 (SD 0.039; Mixed ahead 3/3) & +0.000 (SD 0.003; Mixed ahead 1/3)\tabularnewline
\bottomrule
\end{tabular}}\ifdim\wd\tblbox>\linewidth\resizebox{\linewidth}{!}{\usebox{\tblbox}}\else\usebox{\tblbox}\fi
\endgroup
\caption*{\textbf{Table C.5a:} (columns 2/2) 365M Dense/Mixed pairs under the standard mixed-precision recipe (fp32 master weights and fp32 AdamW state for every parameter; bf16 autocast compute), seeds 41--43, dense operator.}
\end{table*}

\begin{table*}[tp]
\centering
\begingroup\footnotesize\setlength{\tabcolsep}{4pt}
\sbox{\tblbox}{\begin{tabular}{@{}lllll@{}}
\toprule
Seed & Length & Temperatures & $\Delta$ mean $\pm$ 95\% & Mixed better\tabularnewline
\midrule
41 & 24k & default (1.00 / 1.00) & +0.006 $\pm$ 0.005 & 13/36\tabularnewline
41 & 24k & calibrated (0.90 / 0.90) & +0.003 $\pm$ 0.005 & 17/36\tabularnewline
41 & 30k & default (1.00 / 1.00) & $-$0.005 $\pm$ 0.007 & 18/36\tabularnewline
41 & 30k & calibrated (0.90 / 0.85) & $-$0.001 $\pm$ 0.005 & 19/36\tabularnewline
42 & 24k & default (1.00 / 1.00) & +0.012 $\pm$ 0.005 & 9/36\tabularnewline
42 & 24k & calibrated (0.90 / 0.90) & +0.005 $\pm$ 0.004 & 13/36\tabularnewline
42 & 30k & default (1.00 / 1.00) & +0.005 $\pm$ 0.007 & 20/36\tabularnewline
42 & 30k & calibrated (0.90 / 0.85) & +0.002 $\pm$ 0.003 & 17/36\tabularnewline
43 & 24k & default (1.00 / 1.00) & $-$0.027 $\pm$ 0.012 & 27/36\tabularnewline
43 & 24k & calibrated (0.85 / 0.90) & +0.010 $\pm$ 0.010 & 20/36\tabularnewline
43 & 30k & default (1.00 / 1.00) & $-$0.062 $\pm$ 0.016 & 31/36\tabularnewline
43 & 30k & calibrated (0.85 / 0.85) & +0.010 $\pm$ 0.011 & 20/36\tabularnewline
\bottomrule
\end{tabular}}\ifdim\wd\tblbox>\linewidth\resizebox{\linewidth}{!}{\usebox{\tblbox}}\else\usebox{\tblbox}\fi
\endgroup
\caption*{\textbf{Table C.5b:} Standard-recipe pairs on the 36-window set.}
\end{table*}

\begin{table*}[tp]
\centering
\begingroup\scriptsize\setlength{\tabcolsep}{3pt}
\sbox{\tblbox}{\begin{tabular}{@{}lllll@{}}
\toprule
Seed & Recipe & $\tau^{*}$ Dense / Mixed & Dense PPL $\tau$~=~1 $\rightarrow\tau$* (gain) & $\Delta$ default 30k\tabularnewline
\midrule
42 & frozen (canonical) & 0.85 / 0.80 & 55.94 $\rightarrow$ 35.31 (36.9\%) & $-$0.063\tabularnewline
42 & learnable Q/K gains & 0.90 / 0.90 & 33.13 $\rightarrow$ 32.34 (2.4\%) & +0.051\tabularnewline
42 & all 113 gains fp32 & 0.90 / 0.90 & 33.77 $\rightarrow$ 31.85 (5.7\%) & +0.034\tabularnewline
42 & standard recipe (all fp32 master) & 0.90 / 0.85 & 34.55 $\rightarrow$ 31.73 (8.2\%) & $-$0.002\tabularnewline
43 & frozen (canonical) & 0.85 / 0.80 (transferred) & 56.94 $\rightarrow$ 34.87 (38.8\%) & $-$0.047 (diagnostics windows)\tabularnewline
43 & learnable Q/K gains & 0.90 / 0.85 & 34.47 $\rightarrow$ 32.97 (4.4\%) & +0.037\tabularnewline
43 & standard recipe (all fp32 master) & 0.85 / 0.85 & 37.87 $\rightarrow$ 32.29 (14.7\%) & $-$0.070\tabularnewline
\bottomrule
\end{tabular}}\ifdim\wd\tblbox>\linewidth\resizebox{\linewidth}{!}{\usebox{\tblbox}}\else\usebox{\tblbox}\fi
\endgroup
\caption*{\textbf{Table C.5c:} (columns 1/2) Precision ladder at 30k, seeds 42 and 43: frozen gains $\rightarrow$ learnable Q/K gains (56 fp32 tensors) $\rightarrow$ all 113 normalisation gains fp32 (seed 42) $\rightarrow$ standard recipe (all parameters fp32 master). Dense held-out perplexity at $\tau$~=~1 and $\tau^{*}$, Dense calibration gain, Mixed~$-$~Dense $\Delta$ at both temperatures; in-span (8k) $\Delta$ in the last column.}
\end{table*}
\begin{table*}[tp]
\centering
\begingroup\footnotesize\setlength{\tabcolsep}{4pt}
\sbox{\tblbox}{\begin{tabular}{@{}llll@{}}
\toprule
Seed & Recipe & $\Delta$ calibrated 30k & $\Delta$ default 8k\tabularnewline
\midrule
42 & frozen (canonical) & $-$0.016 & +0.008\tabularnewline
42 & learnable Q/K gains & +0.029 & +0.032\tabularnewline
42 & all 113 gains fp32 & +0.023 & +0.021\tabularnewline
42 & standard recipe (all fp32 master) & $-$0.003 & +0.021\tabularnewline
43 & frozen (canonical) & +0.019 (Table C.1c) & +0.029 (diagnostics windows)\tabularnewline
43 & learnable Q/K gains & +0.024 & +0.022\tabularnewline
43 & standard recipe (all fp32 master) & +0.001 & +0.020\tabularnewline
\bottomrule
\end{tabular}}\ifdim\wd\tblbox>\linewidth\resizebox{\linewidth}{!}{\usebox{\tblbox}}\else\usebox{\tblbox}\fi
\endgroup
\caption*{\textbf{Table C.5c:} (columns 2/2) Precision ladder at 30k, seeds 42 and 43: frozen gains $\rightarrow$ learnable Q/K gains (56 fp32 tensors) $\rightarrow$ all 113 normalisation gains fp32 (seed 42) $\rightarrow$ standard recipe (all parameters fp32 master). Dense held-out perplexity at $\tau$~=~1 and $\tau^{*}$, Dense calibration gain, Mixed~$-$~Dense $\Delta$ at both temperatures; in-span (8k) $\Delta$ in the last column.}
\end{table*}

\begin{table*}[tp]
\centering
\begingroup\footnotesize\setlength{\tabcolsep}{4pt}
\sbox{\tblbox}{\begin{tabular}{@{}lllll@{}}
\toprule
Length & $\tau^{*}$ ALiBi / Dense & ALiBi PPL $\tau$~=~1 $\rightarrow\tau$* & Dense PPL $\tau$~=~1 $\rightarrow\tau$* & $\Delta$ default\tabularnewline
\midrule
8k & 0.95 / 1.00 & 35.12 $\rightarrow$ 35.01 & 32.84 $\rightarrow$ 32.84 & +0.067\tabularnewline
16k & 0.95 / 0.90 & 35.47 $\rightarrow$ 35.43 & 36.05 $\rightarrow$ 33.58 & $-$0.016\tabularnewline
24k & 0.95 / 0.85 & 33.61 $\rightarrow$ 33.55 & 42.36 $\rightarrow$ 32.30 & $-$0.232\tabularnewline
30k & 0.95 / 0.85 & 37.01 $\rightarrow$ 36.96 & 55.94 $\rightarrow$ 35.31 & $-$0.413\tabularnewline
\bottomrule
\end{tabular}}\ifdim\wd\tblbox>\linewidth\resizebox{\linewidth}{!}{\usebox{\tblbox}}\else\usebox{\tblbox}\fi
\endgroup
\caption*{\textbf{Table C.6a:} (columns 1/2) The ALiBi seed-42 pair at all four lengths (protocol calibration and held-out windows; pooled held-out perplexity; $\Delta$~=~log PPL(ALiBi) $-$ log PPL(Dense), three-corpus mean). Dense's 16k and 24k cells are the temperature-protocol cells of the supplementary tables; its 8k and 30k cells are the mechanism-controls re-evaluation of Table C.7a (identical values).}
\end{table*}
\begin{table*}[tp]
\centering
\begingroup\footnotesize\setlength{\tabcolsep}{4pt}
\sbox{\tblbox}{\begin{tabular}{@{}llll@{}}
\toprule
Length & $\tau^{*}$ ALiBi / Dense & $\Delta$ calibrated & $\Delta$ calibrated by corpus (arXiv / FineWeb-Edu / PG-19)\tabularnewline
\midrule
8k & 0.95 / 1.00 & +0.064 & +0.116 / +0.007 / +0.070\tabularnewline
16k & 0.95 / 0.90 & +0.054 & +0.140 / $-$0.022 / +0.044\tabularnewline
24k & 0.95 / 0.85 & +0.038 & +0.139 / $-$0.041 / +0.015\tabularnewline
30k & 0.95 / 0.85 & +0.046 & +0.144 / $-$0.041 / +0.034\tabularnewline
\bottomrule
\end{tabular}}\ifdim\wd\tblbox>\linewidth\resizebox{\linewidth}{!}{\usebox{\tblbox}}\else\usebox{\tblbox}\fi
\endgroup
\caption*{\textbf{Table C.6a:} (columns 2/2) The ALiBi seed-42 pair at all four lengths (protocol calibration and held-out windows; pooled held-out perplexity; $\Delta$~=~log PPL(ALiBi) $-$ log PPL(Dense), three-corpus mean). Dense's 16k and 24k cells are the temperature-protocol cells of the supplementary tables; its 8k and 30k cells are the mechanism-controls re-evaluation of Table C.7a (identical values).}
\end{table*}

\begin{table*}[tp]
\centering
\begingroup\scriptsize\setlength{\tabcolsep}{3pt}
\sbox{\tblbox}{\begin{tabular}{@{}llllllll@{}}
\toprule
Seed & Regime & Length & $\tau^{*}$ Dense / ALiBi & Dense PPL $\tau$~=~1 $\rightarrow\tau$* (gain) & ALiBi PPL $\tau$~=~1 $\rightarrow\tau$* (gain) & $\Delta$ default & $\Delta$ calibrated\tabularnewline
\midrule
42 & frozen & 8k & 1.00 / 0.95 & 32.84 $\rightarrow$ 32.84 & 35.12 $\rightarrow$ 35.01 & +0.067 & +0.064\tabularnewline
42 & frozen & 30k & 0.85 / 0.95 & 55.94 $\rightarrow$ 35.31 & 37.01 $\rightarrow$ 36.96 & $-$0.413 & +0.046\tabularnewline
42 & learnable & 8k & 1.00 / 0.95 & 31.32 $\rightarrow$ 31.32 (0.0\%) & 33.31 $\rightarrow$ 33.37 ($-$0.2\%) & +0.062 & +0.063\tabularnewline
42 & learnable & 16k & 0.95 / 0.95 & 31.14 $\rightarrow$ 31.06 (0.3\%) & 34.95 $\rightarrow$ 34.86 (0.3\%) & +0.115 & +0.115\tabularnewline
42 & learnable & 24k & 0.95 / 0.95 & 29.77 $\rightarrow$ 29.32 (1.5\%) & 33.54 $\rightarrow$ 33.44 (0.3\%) & +0.119 & +0.131\tabularnewline
42 & learnable & 30k & 0.90 / 0.95 & 33.13 $\rightarrow$ 32.34 (2.4\%) & 37.11 $\rightarrow$ 37.09 (0.1\%) & +0.113 & +0.137\tabularnewline
43 & frozen & 8k & 1.00 / 0.95 & 32.66 $\rightarrow$ 32.66 & 34.92 $\rightarrow$ 34.80 & +0.067 & +0.063\tabularnewline
43 & frozen & 30k & 0.85 / 0.95 & 56.94 $\rightarrow$ 34.87 & 37.10 $\rightarrow$ 37.04 & $-$0.428 & +0.060\tabularnewline
43 & learnable & 8k & 1.00 / 0.95 & 31.82 $\rightarrow$ 31.82 (0.0\%) & 33.15 $\rightarrow$ 33.22 ($-$0.2\%) & +0.041 & +0.043\tabularnewline
43 & learnable & 16k & 0.95 / 0.95 & 31.80 $\rightarrow$ 31.61 (0.6\%) & 34.53 $\rightarrow$ 34.54 (0.0\%) & +0.082 & +0.089\tabularnewline
43 & learnable & 24k & 0.90 / 0.95 & 30.55 $\rightarrow$ 29.88 (2.2\%) & 32.98 $\rightarrow$ 33.04 ($-$0.2\%) & +0.076 & +0.101\tabularnewline
43 & learnable & 30k & 0.90 / 1.00 & 34.47 $\rightarrow$ 32.97 (4.4\%) & 36.56 $\rightarrow$ 36.56 (0.0\%) & +0.059 & +0.103\tabularnewline
\bottomrule
\end{tabular}}\ifdim\wd\tblbox>\linewidth\resizebox{\linewidth}{!}{\usebox{\tblbox}}\else\usebox{\tblbox}\fi
\endgroup
\caption*{\textbf{Table C.6c:} ALiBi vs RoPE Dense, 365M, seeds 42 and 43: frozen gains (canonical Dense, ALIBI\_20260921 arm) and learnable Q/K gains (FINAL\_20260924 Dense, FINAL7 ALiBi). $\Delta$~=~log PPL(ALiBi) $-$ log PPL(Dense); negative = ALiBi better.}
\end{table*}

\begin{table*}[tp]
\centering
\begingroup\footnotesize\setlength{\tabcolsep}{4pt}
\sbox{\tblbox}{\begin{tabular}{@{}lllll@{}}
\toprule
Seed & Length & Temperatures (Dense / ALiBi) & $\Delta$ mean $\pm$ 95\% & ALiBi better\tabularnewline
\midrule
42 & 24k & default (1.00 / 1.00) & +0.105 $\pm$ 0.029 & 4/36\tabularnewline
42 & 24k & calibrated (0.95 / 0.95) & +0.120 $\pm$ 0.029 & 0/36\tabularnewline
42 & 30k & default (1.00 / 1.00) & +0.092 $\pm$ 0.029 & 11/36\tabularnewline
42 & 30k & calibrated (0.90 / 0.95) & +0.119 $\pm$ 0.030 & 0/36\tabularnewline
43 & 24k & default (1.00 / 1.00) & +0.065 $\pm$ 0.024 & 12/36\tabularnewline
43 & 24k & calibrated (0.90 / 0.95) & +0.090 $\pm$ 0.024 & 1/36\tabularnewline
43 & 30k & default (1.00 / 1.00) & +0.040 $\pm$ 0.025 & 13/36\tabularnewline
43 & 30k & calibrated (0.90 / 1.00) & +0.086 $\pm$ 0.025 & 8/36\tabularnewline
\bottomrule
\end{tabular}}\ifdim\wd\tblbox>\linewidth\resizebox{\linewidth}{!}{\usebox{\tblbox}}\else\usebox{\tblbox}\fi
\endgroup
\caption*{\textbf{Table C.6d:} ALiBi (learnable gains) vs learnable-gain Dense on the 36-window set.}
\end{table*}

\begin{table*}[tp]
\centering
\begingroup\footnotesize\setlength{\tabcolsep}{4pt}
\sbox{\tblbox}{\begin{tabular}{@{}llllll@{}}
\toprule
Control arm (365M) & Length & Seeds & Control $\tau^{*}$ & Dense $\tau^{*}$ & Mean $\Delta$PPL at $\tau$~=~1\tabularnewline
\midrule
Static fixed mask (mean budget) & 8k & 6 & 0.95 & 1.00 & +0.5070\tabularnewline
Static fixed mask (mean budget) & 30k & 6 & 0.80 & 0.85 & +20.3603\tabularnewline
Resampled fixed-budget mask (mean budget) & 8k & 6 & 0.95 & 1.00 & +0.5275\tabularnewline
Resampled fixed-budget mask (mean budget) & 30k & 6 & 0.80 & 0.85 & +16.9479\tabularnewline
Attention dropout p = 0.28 & 8k & 5 & 0.95/1.00 & 1.00 & +0.3774\tabularnewline
Attention dropout p = 0.28 & 30k & 5 & 0.80/0.85 & 0.85 & $-$4.5420\tabularnewline
RoPE-distance control & 8k & 5 & 1.00 & 1.00 & +0.0453\tabularnewline
RoPE-distance control & 30k & 5 & 0.85 & 0.85 & +3.1360\tabularnewline
ALiBi (no RoPE) & 8k & 5 & 0.95 & 1.00 & +2.1911\tabularnewline
ALiBi (no RoPE) & 30k & 5 & 0.95 & 0.85 & $-$19.5560\tabularnewline
\bottomrule
\end{tabular}}\ifdim\wd\tblbox>\linewidth\resizebox{\linewidth}{!}{\usebox{\tblbox}}\else\usebox{\tblbox}\fi
\endgroup
\caption*{\textbf{Table C.7a:} (columns 1/2) The 365M control arms under the held-out temperature protocol (frozen-gain checkpoints; dense operator; pooled held-out perplexity over arXiv, FineWeb-Edu and PG-19; $\Delta$~=~control $-$ Dense in perplexity points, so negative favours the control). Each model selects its own $\tau$ on the calibration split (13-point grid) and is scored on the held-out split at $\tau$~=~1 and at its $\tau^{*}$.}
\end{table*}
\begin{table*}[tp]
\centering
\begingroup\scriptsize\setlength{\tabcolsep}{3pt}
\sbox{\tblbox}{\begin{tabular}{@{}lllll@{}}
\toprule
Control arm (365M) & Length & Control better at $\tau$~=~1 & Mean $\Delta$PPL at own $\tau^{*}$ & Control better calibrated\tabularnewline
\midrule
Static fixed mask (mean budget) & 8k & 0/6 & +0.0910 & 2/6\tabularnewline
Static fixed mask (mean budget) & 30k & 0/6 & $-$0.2371 & 4/6\tabularnewline
Resampled fixed-budget mask (mean budget) & 8k & 0/6 & +0.1693 & 1/6\tabularnewline
Resampled fixed-budget mask (mean budget) & 30k & 0/6 & $-$0.2916 & 5/6\tabularnewline
Attention dropout p = 0.28 & 8k & 1/5 & +0.3873 & 1/5\tabularnewline
Attention dropout p = 0.28 & 30k & 5/5 & $-$0.3040 & 4/5\tabularnewline
RoPE-distance control & 8k & 2/5 & +0.0453 & 2/5\tabularnewline
RoPE-distance control & 30k & 0/5 & +0.0216 & 1/5\tabularnewline
ALiBi (no RoPE) & 8k & 0/5 & +2.0816 & 0/5\tabularnewline
ALiBi (no RoPE) & 30k & 5/5 & +1.6321 & 0/5\tabularnewline
\bottomrule
\end{tabular}}\ifdim\wd\tblbox>\linewidth\resizebox{\linewidth}{!}{\usebox{\tblbox}}\else\usebox{\tblbox}\fi
\endgroup
\caption*{\textbf{Table C.7a:} (columns 2/2) The 365M control arms under the held-out temperature protocol (frozen-gain checkpoints; dense operator; pooled held-out perplexity over arXiv, FineWeb-Edu and PG-19; $\Delta$~=~control $-$ Dense in perplexity points, so negative favours the control). Each model selects its own $\tau$ on the calibration split (13-point grid) and is scored on the held-out split at $\tau$~=~1 and at its $\tau^{*}$.}
\end{table*}

\begin{table*}[tp]
\centering
\begingroup\scriptsize\setlength{\tabcolsep}{3pt}
\sbox{\tblbox}{\begin{tabular}{@{}lllllll@{}}
\toprule
Pair & Length & $\tau^{*}$ Dense / Mixed & Dense PPL $\tau$~=~1 $\rightarrow\tau$* & Mixed PPL $\tau$~=~1 $\rightarrow\tau$* & $\Delta$ default & $\Delta$ calibrated\tabularnewline
\midrule
no QK-norm (base \(10^{6}\)) & 8k & 1.00 / 1.00 & 32.00 $\rightarrow$ 32.00 & 32.73 $\rightarrow$ 32.73 & +0.022 & +0.022\tabularnewline
no QK-norm (base \(10^{6}\)) & 16k & 0.85 / 0.80 & 44.90 $\rightarrow$ 42.16 & 48.62 $\rightarrow$ 45.03 & +0.080 & +0.066\tabularnewline
no QK-norm (base \(10^{6}\)) & 24k & 0.75 / 0.70 & 75.06 $\rightarrow$ 61.67 & 85.24 $\rightarrow$ 72.23 & +0.127 & +0.158\tabularnewline
no QK-norm (base \(10^{6}\)) & 30k & 0.75 / 0.70 & 107.96 $\rightarrow$ 82.91 & 122.59 $\rightarrow$ 100.92 & +0.127 & +0.197\tabularnewline
RoPE base 10$^{4}$ (QK-norm) & 8k & 1.00 / 0.95 & 33.13 $\rightarrow$ 33.13 & 33.74 $\rightarrow$ 33.54 & +0.018 & +0.012\tabularnewline
RoPE base 10$^{4}$ (QK-norm) & 16k & 0.90 / 0.85 & 37.71 $\rightarrow$ 34.92 & 38.50 $\rightarrow$ 34.58 & +0.021 & $-$0.010\tabularnewline
RoPE base 10$^{4}$ (QK-norm) & 24k & 0.90 / 0.85 & 46.49 $\rightarrow$ 34.82 & 44.66 $\rightarrow$ 33.49 & $-$0.040 & $-$0.039\tabularnewline
RoPE base 10$^{4}$ (QK-norm) & 30k & 0.85 / 0.80 & 61.36 $\rightarrow$ 39.26 & 58.33 $\rightarrow$ 37.32 & $-$0.051 & $-$0.051\tabularnewline
\bottomrule
\end{tabular}}\ifdim\wd\tblbox>\linewidth\resizebox{\linewidth}{!}{\usebox{\tblbox}}\else\usebox{\tblbox}\fi
\endgroup
\caption*{\textbf{Table C.8a:} The regime pairs at all four lengths (protocol calibration and held-out windows; pooled held-out perplexity; $\Delta$~=~log PPL(Mixed) $-$ log PPL(Dense), three-corpus mean).}
\end{table*}

\begin{table*}[tp]
\centering
\begingroup\scriptsize\setlength{\tabcolsep}{3pt}
\sbox{\tblbox}{\begin{tabular}{@{}lllll@{}}
\toprule
Regime (365M, seed 42) & Length & $\tau^{*}$ Dense / Mixed & Dense PPL $\tau$~=~1 $\rightarrow\tau$* (gain) & Mixed PPL $\tau$~=~1 $\rightarrow\tau$* (gain)\tabularnewline
\midrule
frozen (canonical) & 8k & 1.00 / 0.95 & 32.8 $\rightarrow$ 32.8 (0.0\%) & 33.1 $\rightarrow$ 32.9 (0.7\%)\tabularnewline
frozen (canonical) & 16k & 0.90 / 0.90 & 36.0 $\rightarrow$ 33.6 (6.8\%) & 36.3 $\rightarrow$ 33.1 (8.7\%)\tabularnewline
frozen (canonical) & 24k & 0.85 / 0.85 & 42.4 $\rightarrow$ 32.3 (23.8\%) & 40.5 $\rightarrow$ 31.5 (22.4\%)\tabularnewline
frozen (canonical) & 30k & 0.85 / 0.80 & 55.9 $\rightarrow$ 35.3 (36.9\%) & 52.5 $\rightarrow$ 34.8 (33.9\%)\tabularnewline
scalar Q/K gains & 8k & 1.00 / 1.00 & 32.4 $\rightarrow$ 32.4 (0.0\%) & 32.1 $\rightarrow$ 32.1 (0.0\%)\tabularnewline
scalar Q/K gains & 16k & 1.00 / 0.95 & 32.0 $\rightarrow$ 32.0 (0.0\%) & 31.9 $\rightarrow$ 31.8 (0.3\%)\tabularnewline
scalar Q/K gains & 24k & 0.95 / 0.95 & 30.2 $\rightarrow$ 30.2 (0.1\%) & 30.3 $\rightarrow$ 30.0 (0.9\%)\tabularnewline
scalar Q/K gains & 30k & 0.95 / 0.95 & 33.3 $\rightarrow$ 33.2 (0.4\%) & 33.4 $\rightarrow$ 33.0 (1.3\%)\tabularnewline
vector Q/K gains (FINAL\_20260924, seed 42) & 8k & 1.00 / 0.95 & 31.3 $\rightarrow$ 31.3 (0.0\%) & 32.3 $\rightarrow$ 32.3 ($-$0.1\%)\tabularnewline
vector Q/K gains (FINAL\_20260924, seed 42) & 16k & 0.95 / 0.95 & 31.1 $\rightarrow$ 31.1 (0.3\%) & 32.4 $\rightarrow$ 32.1 (1.0\%)\tabularnewline
vector Q/K gains (FINAL\_20260924, seed 42) & 24k & 0.95 / 0.90 & 29.8 $\rightarrow$ 29.3 (1.5\%) & 31.2 $\rightarrow$ 30.2 (3.2\%)\tabularnewline
vector Q/K gains (FINAL\_20260924, seed 42) & 30k & 0.90 / 0.90 & 33.1 $\rightarrow$ 32.3 (2.4\%) & 34.9 $\rightarrow$ 33.3 (4.6\%)\tabularnewline
all 113 gains fp32 & 8k & 1.00 / 0.95 & 30.5 $\rightarrow$ 30.5 (0.0\%) & 31.1 $\rightarrow$ 31.2 ($-$0.1\%)\tabularnewline
all 113 gains fp32 & 16k & 0.95 / 0.95 & 30.7 $\rightarrow$ 30.4 (0.9\%) & 31.6 $\rightarrow$ 31.1 (1.4\%)\tabularnewline
all 113 gains fp32 & 24k & 0.90 / 0.90 & 29.9 $\rightarrow$ 28.9 (3.2\%) & 30.9 $\rightarrow$ 29.6 (4.3\%)\tabularnewline
all 113 gains fp32 & 30k & 0.90 / 0.90 & 33.8 $\rightarrow$ 31.8 (5.7\%) & 34.9 $\rightarrow$ 32.6 (6.7\%)\tabularnewline
 &  &  &  & \tabularnewline
RoPE 10$^{4}$, frozen gains (REGIME\_20260922) & 8k & 1.00 / 0.95 & 33.1 $\rightarrow$ 33.1 (0.0\%) & 33.7 $\rightarrow$ 33.5 (0.6\%)\tabularnewline
RoPE 10$^{4}$, frozen gains (REGIME\_20260922) & 16k & 0.90 / 0.85 & 37.7 $\rightarrow$ 34.9 (7.4\%) & 38.5 $\rightarrow$ 34.6 (10.2\%)\tabularnewline
RoPE 10$^{4}$, frozen gains (REGIME\_20260922) & 24k & 0.90 / 0.85 & 46.5 $\rightarrow$ 34.8 (25.1\%) & 44.7 $\rightarrow$ 33.5 (25.0\%)\tabularnewline
RoPE 10$^{4}$, frozen gains (REGIME\_20260922) & 30k & 0.85 / 0.80 & 61.4 $\rightarrow$ 39.3 (36.0\%) & 58.3 $\rightarrow$ 37.3 (36.0\%)\tabularnewline
RoPE 10$^{4}$, learnable Q/K gains & 8k & 1.00 / 0.95 & 32.1 $\rightarrow$ 32.1 (0.0\%) & 32.9 $\rightarrow$ 32.9 ($-$0.2\%)\tabularnewline
RoPE 10$^{4}$, learnable Q/K gains & 16k & 0.95 / 0.95 & 32.2 $\rightarrow$ 32.1 (0.2\%) & 32.9 $\rightarrow$ 32.8 (0.5\%)\tabularnewline
RoPE 10$^{4}$, learnable Q/K gains & 24k & 0.95 / 0.90 & 31.2 $\rightarrow$ 30.7 (1.5\%) & 31.9 $\rightarrow$ 31.4 (1.6\%)\tabularnewline
RoPE 10$^{4}$, learnable Q/K gains & 30k & 0.90 / 0.90 & 35.5 $\rightarrow$ 34.6 (2.6\%) & 36.1 $\rightarrow$ 35.1 (2.9\%)\tabularnewline
\bottomrule
\end{tabular}}\ifdim\wd\tblbox>\linewidth\resizebox{\linewidth}{!}{\usebox{\tblbox}}\else\usebox{\tblbox}\fi
\endgroup
\caption*{\textbf{Table C.8c:} (columns 1/2) The seed-42 365M regimes: frozen (canonical) gains, scalar Q/K gains, vector Q/K gains (\path{FINAL_20260924}) and all 113 gains in fp32, and the RoPE-10$^{4}$ pair with frozen and with learnable gains. The frozen RoPE-10$^{4}$ pair's 36-window residuals are $-$0.068 $\pm$ 0.028 (default) and $-$0.054 $\pm$ 0.010 (calibrated) at 30k with Mixed ahead in 26 and 36 of 36 (the supplementary tables); with learnable gains +0.024 $\pm$ 0.006 and +0.018 $\pm$ 0.005, Mixed ahead in 3 and 2 of 36 (supplementary tables).}
\end{table*}
\begin{table*}[tp]
\centering
\begingroup\footnotesize\setlength{\tabcolsep}{4pt}
\sbox{\tblbox}{\begin{tabular}{@{}llll@{}}
\toprule
Regime (365M, seed 42) & Length & $\Delta$ default & $\Delta$ calibrated\tabularnewline
\midrule
frozen (canonical) & 8k & +0.008 & +0.000\tabularnewline
frozen (canonical) & 16k & +0.006 & $-$0.014\tabularnewline
frozen (canonical) & 24k & $-$0.044 & $-$0.026\tabularnewline
frozen (canonical) & 30k & $-$0.063 & $-$0.016\tabularnewline
scalar Q/K gains & 8k & $-$0.008 & $-$0.008\tabularnewline
scalar Q/K gains & 16k & $-$0.003 & $-$0.006\tabularnewline
scalar Q/K gains & 24k & +0.004 & $-$0.005\tabularnewline
scalar Q/K gains & 30k & +0.005 & $-$0.005\tabularnewline
vector Q/K gains (FINAL\_20260924, seed 42) & 8k & +0.032 & +0.032\tabularnewline
vector Q/K gains (FINAL\_20260924, seed 42) & 16k & +0.040 & +0.033\tabularnewline
vector Q/K gains (FINAL\_20260924, seed 42) & 24k & +0.047 & +0.031\tabularnewline
vector Q/K gains (FINAL\_20260924, seed 42) & 30k & +0.051 & +0.029\tabularnewline
all 113 gains fp32 & 8k & +0.021 & +0.023\tabularnewline
all 113 gains fp32 & 16k & +0.029 & +0.024\tabularnewline
all 113 gains fp32 & 24k & +0.035 & +0.024\tabularnewline
all 113 gains fp32 & 30k & +0.034 & +0.023\tabularnewline
 &  &  & \tabularnewline
RoPE 10$^{4}$, frozen gains (REGIME\_20260922) & 8k & +0.018 & +0.012\tabularnewline
RoPE 10$^{4}$, frozen gains (REGIME\_20260922) & 16k & +0.021 & $-$0.010\tabularnewline
RoPE 10$^{4}$, frozen gains (REGIME\_20260922) & 24k & $-$0.040 & $-$0.039\tabularnewline
RoPE 10$^{4}$, frozen gains (REGIME\_20260922) & 30k & $-$0.051 & $-$0.051\tabularnewline
RoPE 10$^{4}$, learnable Q/K gains & 8k & +0.024 & +0.025\tabularnewline
RoPE 10$^{4}$, learnable Q/K gains & 16k & +0.023 & +0.020\tabularnewline
RoPE 10$^{4}$, learnable Q/K gains & 24k & +0.023 & +0.022\tabularnewline
RoPE 10$^{4}$, learnable Q/K gains & 30k & +0.016 & +0.013\tabularnewline
\bottomrule
\end{tabular}}\ifdim\wd\tblbox>\linewidth\resizebox{\linewidth}{!}{\usebox{\tblbox}}\else\usebox{\tblbox}\fi
\endgroup
\caption*{\textbf{Table C.8c:} (columns 2/2) The seed-42 365M regimes: frozen (canonical) gains, scalar Q/K gains, vector Q/K gains (\path{FINAL_20260924}) and all 113 gains in fp32, and the RoPE-10$^{4}$ pair with frozen and with learnable gains. The frozen RoPE-10$^{4}$ pair's 36-window residuals are $-$0.068 $\pm$ 0.028 (default) and $-$0.054 $\pm$ 0.010 (calibrated) at 30k with Mixed ahead in 26 and 36 of 36 (the supplementary tables); with learnable gains +0.024 $\pm$ 0.006 and +0.018 $\pm$ 0.005, Mixed ahead in 3 and 2 of 36 (supplementary tables).}
\end{table*}

\begin{table*}[tp]
\centering
\begingroup\scriptsize\setlength{\tabcolsep}{3pt}
\sbox{\tblbox}{\begin{tabular}{@{}lllllll@{}}
\toprule
Pair & Length & $\tau^{*}$ A / B & A PPL $\tau$~=~1 $\rightarrow\tau$* (gain) & B PPL $\tau$~=~1 $\rightarrow\tau$* (gain) & $\Delta$ default & $\Delta$ calibrated\tabularnewline
\midrule
1.2B s70, learnable Q/K & 8k & 1.00 / 0.95 & 21.70 $\rightarrow$ 21.70 (0.0\%) & 21.83 $\rightarrow$ 21.85 ($-$0.1\%) & +0.006 & +0.007\tabularnewline
1.2B s70, learnable Q/K & 16k & 0.95 / 0.95 & 21.79 $\rightarrow$ 21.51 (1.3\%) & 21.75 $\rightarrow$ 21.60 (0.7\%) & $-$0.002 & +0.004\tabularnewline
1.2B s70, learnable Q/K & 24k & 0.90 / 0.90 & 21.72 $\rightarrow$ 20.28 (6.7\%) & 20.74 $\rightarrow$ 20.34 (1.9\%) & $-$0.046 & +0.003\tabularnewline
1.2B s70, learnable Q/K & 30k & 0.90 / 0.90 & 25.23 $\rightarrow$ 22.41 (11.2\%) & 23.04 $\rightarrow$ 22.39 (2.8\%) & $-$0.091 & $-$0.001\tabularnewline
365M s42, all-norm fp32 & 8k & 1.00 / 0.95 & 30.48 $\rightarrow$ 30.48 (0.0\%) & 31.14 $\rightarrow$ 31.18 ($-$0.1\%) & +0.021 & +0.023\tabularnewline
365M s42, all-norm fp32 & 16k & 0.95 / 0.95 & 30.68 $\rightarrow$ 30.40 (0.9\%) & 31.59 $\rightarrow$ 31.13 (1.4\%) & +0.029 & +0.024\tabularnewline
365M s42, all-norm fp32 & 24k & 0.90 / 0.90 & 29.86 $\rightarrow$ 28.89 (3.2\%) & 30.91 $\rightarrow$ 29.58 (4.3\%) & +0.035 & +0.024\tabularnewline
365M s42, all-norm fp32 & 30k & 0.90 / 0.90 & 33.77 $\rightarrow$ 31.85 (5.7\%) & 34.94 $\rightarrow$ 32.60 (6.7\%) & +0.034 & +0.023\tabularnewline
365M s42, scalar Q/K & 8k & 1.00 / 1.00 & 32.41 $\rightarrow$ 32.41 (0.0\%) & 32.15 $\rightarrow$ 32.15 (0.0\%) & $-$0.008 & $-$0.008\tabularnewline
365M s42, scalar Q/K & 16k & 1.00 / 0.95 & 31.96 $\rightarrow$ 31.96 (0.0\%) & 31.86 $\rightarrow$ 31.77 (0.3\%) & $-$0.003 & $-$0.006\tabularnewline
365M s42, scalar Q/K & 24k & 0.95 / 0.95 & 30.17 $\rightarrow$ 30.15 (0.1\%) & 30.29 $\rightarrow$ 30.01 (0.9\%) & +0.004 & $-$0.005\tabularnewline
365M s42, scalar Q/K & 30k & 0.95 / 0.95 & 33.27 $\rightarrow$ 33.16 (0.4\%) & 33.44 $\rightarrow$ 32.99 (1.3\%) & +0.005 & $-$0.005\tabularnewline
365M s42, RoPE 10$^{4}$ + learnable Q/K & 8k & 1.00 / 0.95 & 32.12 $\rightarrow$ 32.12 (0.0\%) & 32.90 $\rightarrow$ 32.95 ($-$0.2\%) & +0.024 & +0.025\tabularnewline
365M s42, RoPE 10$^{4}$ + learnable Q/K & 16k & 0.95 / 0.95 & 32.20 $\rightarrow$ 32.13 (0.2\%) & 32.94 $\rightarrow$ 32.78 (0.5\%) & +0.023 & +0.020\tabularnewline
365M s42, RoPE 10$^{4}$ + learnable Q/K & 24k & 0.95 / 0.90 & 31.19 $\rightarrow$ 30.71 (1.5\%) & 31.90 $\rightarrow$ 31.40 (1.6\%) & +0.023 & +0.022\tabularnewline
365M s42, RoPE 10$^{4}$ + learnable Q/K & 30k & 0.90 / 0.90 & 35.51 $\rightarrow$ 34.61 (2.6\%) & 36.09 $\rightarrow$ 35.05 (2.9\%) & +0.016 & +0.013\tabularnewline
365M s42, Dense-16k (B) vs Dense-8k (A) & 8k & 1.00 / 1.00 & 31.32 $\rightarrow$ 31.32 (0.0\%) & 32.22 $\rightarrow$ 32.22 (0.0\%) & +0.028 & +0.028\tabularnewline
365M s42, Dense-16k (B) vs Dense-8k (A) & 16k & 0.95 / 1.00 & 31.14 $\rightarrow$ 31.06 (0.3\%) & 31.74 $\rightarrow$ 31.74 (0.0\%) & +0.019 & +0.022\tabularnewline
365M s42, Dense-16k (B) vs Dense-8k (A) & 24k & 0.95 / 1.00 & 29.78 $\rightarrow$ 29.32 (1.5\%) & 29.90 $\rightarrow$ 29.90 (0.0\%) & +0.004 & +0.019\tabularnewline
365M s42, Dense-16k (B) vs Dense-8k (A) & 30k & 0.90 / 0.95 & 33.13 $\rightarrow$ 32.34 (2.4\%) & 33.27 $\rightarrow$ 33.25 (0.1\%) & +0.004 & +0.028\tabularnewline
\bottomrule
\end{tabular}}\ifdim\wd\tblbox>\linewidth\resizebox{\linewidth}{!}{\usebox{\tblbox}}\else\usebox{\tblbox}\fi
\endgroup
\caption*{\textbf{Table C.8d:} The five later pairs under the protocol at 8k--30k (held-out windows). A = Dense (for the last pair the 8k twin); B = Mixed (for the last pair the 16k model); s70/s42 = seed. Gains in parentheses: the arm's own perplexity reduction from $\tau$~=~1 to its $\tau^{*}$.}
\end{table*}

\begin{table*}[tp]
\centering
\begingroup\footnotesize\setlength{\tabcolsep}{4pt}
\sbox{\tblbox}{\begin{tabular}{@{}lllllll@{}}
\toprule
Tier & Seed & Context & $\Delta$ at $\tau$=0.80 (both arms) & $\Delta$ at $\tau$=0.85 (both arms) & $\Delta$ at $\tau$=0.90 (both arms) & $\Delta$ at $\tau$=1.00 (both arms)\tabularnewline
\midrule
365M & 42 & 8k & $-$0.0850 & $-$0.0527 & $-$0.0250 & 0.0094\tabularnewline
365M & 42 & 16k & $-$0.0726 & $-$0.0403 & $-$0.0148 & 0.0003\tabularnewline
365M & 42 & 24k & $-$0.0513 & $-$0.0227 & $-$0.0030 & $-$0.0574\tabularnewline
365M & 42 & 30k & $-$0.0399 & $-$0.0112 & 0.0063 & $-$0.0819\tabularnewline
1.2B & 42 & 8k & $-$0.0902 & $-$0.0409 & $-$0.0106 & 0.0148\tabularnewline
1.2B & 42 & 16k & $-$0.0703 & $-$0.0301 & $-$0.0045 & 0.0054\tabularnewline
1.2B & 42 & 24k & $-$0.0458 & $-$0.0130 & 0.0085 & $-$0.0848\tabularnewline
1.2B & 42 & 30k & $-$0.0324 & $-$0.0020 & 0.0144 & $-$0.1806\tabularnewline
1.2B & 71 & 8k & $-$0.0379 & $-$0.0248 & $-$0.0126 & 0.0052\tabularnewline
1.2B & 71 & 30k & $-$0.0191 & $-$0.0013 & 0.0099 & $-$0.2217\tabularnewline
3B & 42 & 8k & $-$0.0516 & $-$0.0318 & $-$0.0133 & 0.0094\tabularnewline
3B & 42 & 16k & $-$0.0491 & $-$0.0273 & $-$0.0097 & $-$0.0259\tabularnewline
3B & 42 & 24k & $-$0.0305 & $-$0.0094 & 0.0024 & $-$0.1922\tabularnewline
3B & 42 & 30k & $-$0.0113 & 0.0182 & 0.0244 & $-$0.3407\tabularnewline
\bottomrule
\end{tabular}}\ifdim\wd\tblbox>\linewidth\resizebox{\linewidth}{!}{\usebox{\tblbox}}\else\usebox{\tblbox}\fi
\endgroup
\caption*{\textbf{Table C.9b:} Both arms at a common temperature on the calibration windows (dense operator; $\Delta$~=~mean NLL difference over three corpora). At $\tau$~=~0.85 (Dense's 30k optimum) the 30k gap is near zero at 365M and 1.2B and slightly positive at 3B; at $\tau$~=~0.80 (Mixed's optimum) Mixed leads by 0.01--0.04.}
\end{table*}

\begin{table*}[tp]
\centering
\begingroup\footnotesize\setlength{\tabcolsep}{4pt}
\sbox{\tblbox}{\begin{tabular}{@{}llllll@{}}
\toprule
Model & Length & Baseline $\tau^{*}$ & Baseline PPL $\tau$~=~1 & Baseline PPL $\tau^{*}$ & NTK $\tau^{*}$\tabularnewline
\midrule
365M Dense-42 & 16k & 0.90 & 36.048 & 33.579 & 0.90\tabularnewline
365M Dense-42 & 24k & 0.85 & 42.362 & 32.298 & 0.85\tabularnewline
365M Dense-42 & 30k & 0.85 & 55.940 & 35.311 & 0.85\tabularnewline
1.2B Dense-42 & 16k & 0.90 & 24.248 & 22.549 & 0.90\tabularnewline
1.2B Dense-42 & 24k & 0.90 & 31.411 & 21.662 & 0.85\tabularnewline
1.2B Dense-42 & 30k & 0.85 & 47.806 & 23.994 & 0.85\tabularnewline
3B Dense-42 & 16k & 0.90 & 20.250 & 18.983 & 0.90\tabularnewline
3B Dense-42 & 24k & 0.85 & 27.452 & 18.109 & 0.85\tabularnewline
3B Dense-42 & 30k & 0.85 & 47.213 & 20.203 & 0.85\tabularnewline
\bottomrule
\end{tabular}}\ifdim\wd\tblbox>\linewidth\resizebox{\linewidth}{!}{\usebox{\tblbox}}\else\usebox{\tblbox}\fi
\endgroup
\caption*{\textbf{Table C.9c:} (columns 1/2) NTK-aware RoPE scaling at inference as a positive control for the calibration protocol (frozen-gain checkpoints; Dense seed 42 of each tier; pooled held-out perplexity over the three corpora). For extension factor $\alpha$ = evaluation length / 8,192 the rotary base is set to base $\times \alpha^{d/(d-2)}$ at inference only (the 8k case is the identity); no training was performed.}
\end{table*}
\begin{table*}[tp]
\centering
\begingroup\footnotesize\setlength{\tabcolsep}{4pt}
\sbox{\tblbox}{\begin{tabular}{@{}llllll@{}}
\toprule
Model & Length & NTK PPL $\tau$~=~1 & NTK PPL $\tau^{*}$ & Gain $\tau$~=~1 (\%) & Gain $\tau^{*}$ (\%)\tabularnewline
\midrule
365M Dense-42 & 16k & 34.449 & 32.562 & 4.4 & 3.0\tabularnewline
365M Dense-42 & 24k & 39.540 & 31.128 & 6.7 & 3.6\tabularnewline
365M Dense-42 & 30k & 51.495 & 34.118 & 7.9 & 3.4\tabularnewline
1.2B Dense-42 & 16k & 22.972 & 21.786 & 5.3 & 3.4\tabularnewline
1.2B Dense-42 & 24k & 26.839 & 20.691 & 14.6 & 4.5\tabularnewline
1.2B Dense-42 & 30k & 37.201 & 22.923 & 22.2 & 4.5\tabularnewline
3B Dense-42 & 16k & 19.069 & 18.410 & 5.8 & 3.0\tabularnewline
3B Dense-42 & 24k & 21.378 & 17.364 & 22.1 & 4.1\tabularnewline
3B Dense-42 & 30k & 29.408 & 19.303 & 37.7 & 4.5\tabularnewline
\bottomrule
\end{tabular}}\ifdim\wd\tblbox>\linewidth\resizebox{\linewidth}{!}{\usebox{\tblbox}}\else\usebox{\tblbox}\fi
\endgroup
\caption*{\textbf{Table C.9c:} (columns 2/2) NTK-aware RoPE scaling at inference as a positive control for the calibration protocol (frozen-gain checkpoints; Dense seed 42 of each tier; pooled held-out perplexity over the three corpora). For extension factor $\alpha$ = evaluation length / 8,192 the rotary base is set to base $\times \alpha^{d/(d-2)}$ at inference only (the 8k case is the identity); no training was performed.}
\end{table*}

\begin{table*}[tp]
\centering
\begingroup\footnotesize\setlength{\tabcolsep}{4pt}
\sbox{\tblbox}{\begin{tabular}{@{}lllllll@{}}
\toprule
Pair & Length & Rule & $\tau$ & Dense pooled PPL & Mixed pooled PPL & $\Delta$ corpus-mean log-PPL\tabularnewline
\midrule
365M-42 & 16k & YaRN (0.1 ln s + 1)$^{-2}$ & 0.87 & 25.43 & 24.79 & $-$0.025\tabularnewline
365M-42 & 16k & ln(8192)/ln(L) & 0.929 & 25.08 & 25.04 & $-$0.002\tabularnewline
365M-42 & 24k & ln(8192)/ln(L) & 0.891 & 32.05 & 31.79 & $-$0.008\tabularnewline
365M-42 & 30k & YaRN (0.1 ln s + 1)$^{-2}$ & 0.78 & 37.33 & 35.01 & $-$0.064\tabularnewline
365M-42 & 30k & ln(8192)/ln(L) & 0.872 & 35.37 & 35.15 & $-$0.006\tabularnewline
1.2B-71 & 16k & YaRN (0.1 ln s + 1)$^{-2}$ & 0.87 & 16.72 & 16.39 & $-$0.020\tabularnewline
1.2B-71 & 16k & ln(8192)/ln(L) & 0.929 & 16.54 & 16.39 & $-$0.009\tabularnewline
1.2B-71 & 24k & YaRN (0.1 ln s + 1)$^{-2}$ & 0.81 & 22.01 & 21.47 & $-$0.025\tabularnewline
1.2B-71 & 24k & ln(8192)/ln(L) & 0.891 & 21.49 & 21.42 & $-$0.003\tabularnewline
1.2B-71 & 30k & YaRN (0.1 ln s + 1)$^{-2}$ & 0.78 & 24.85 & 24.15 & $-$0.028\tabularnewline
1.2B-71 & 30k & ln(8192)/ln(L) & 0.872 & 24.01 & 24.12 & +0.005\tabularnewline
3B-42 & 16k & YaRN (0.1 ln s + 1)$^{-2}$ & 0.87 & 13.81 & 13.48 & $-$0.024\tabularnewline
3B-42 & 16k & ln(8192)/ln(L) & 0.929 & 13.64 & 13.49 & $-$0.011\tabularnewline
3B-42 & 24k & YaRN (0.1 ln s + 1)$^{-2}$ & 0.81 & 18.51 & 17.64 & $-$0.048\tabularnewline
3B-42 & 24k & ln(8192)/ln(L) & 0.891 & 18.11 & 17.79 & $-$0.018\tabularnewline
3B-42 & 30k & YaRN (0.1 ln s + 1)$^{-2}$ & 0.78 & 20.90 & 20.31 & $-$0.029\tabularnewline
3B-42 & 30k & ln(8192)/ln(L) & 0.872 & 20.39 & 20.87 & +0.023\tabularnewline
\bottomrule
\end{tabular}}\ifdim\wd\tblbox>\linewidth\resizebox{\linewidth}{!}{\usebox{\tblbox}}\else\usebox{\tblbox}\fi
\endgroup
\caption*{\textbf{Table C.10a:} Data-free rule temperatures applied to both arms on the held-out windows (frozen-gain checkpoints; 365M seed 42, 1.2B seed 71 and 3B seed 42; dense operator; the held-out windows of Table C.1c: two per corpus; at 16k these are not the four-window held-out set of Appendix B, so 16k perplexities are not comparable with Tables C.8c and C.9c). Two rules that need no calibration data: YaRN's $\tau$~=~(0.1 ln s + 1)$^{-2}$ with s = L/8,192 (0.87, 0.81, 0.78 at 16k, 24k, 30k) and the entropy-invariance rule $\tau$~=~ln(8,192)/ln(L) (0.929, 0.891, 0.872). $\Delta$~=~corpus-mean log PPL(Mixed) $-$ log PPL(Dense) with both arms at the same $\tau$.}
\end{table*}

\begin{table*}[tp]
\centering
\begingroup\scriptsize\setlength{\tabcolsep}{3pt}
\sbox{\tblbox}{\begin{tabular}{@{}lllllll@{}}
\toprule
Layer & Dense \(g_q\) & Dense \(g_k\) & Dense \(g_q g_k\) & Mixed \(g_q\) & Mixed \(g_k\) & Mixed \(g_q g_k\)\tabularnewline
\midrule
0 & 0.44 & 0.45 & 0.24 & 0.42 & 0.43 & 0.23\tabularnewline
1 & 1.04 & 1.04 & 1.10 & 1.18 & 1.19 & 1.41\tabularnewline
2 & 1.21 & 1.20 & 1.47 & 1.30 & 1.29 & 1.69\tabularnewline
3 & 1.32 & 1.31 & 1.75 & 1.34 & 1.33 & 1.80\tabularnewline
4 & 1.22 & 1.22 & 1.50 & 1.32 & 1.32 & 1.75\tabularnewline
5 & 1.27 & 1.26 & 1.61 & 1.39 & 1.39 & 1.96\tabularnewline
6 & 1.24 & 1.22 & 1.58 & 1.42 & 1.40 & 2.07\tabularnewline
7 & 1.36 & 1.34 & 1.86 & 1.41 & 1.40 & 2.01\tabularnewline
8 & 1.36 & 1.34 & 1.86 & 1.47 & 1.45 & 2.18\tabularnewline
9 & 1.41 & 1.39 & 1.98 & 1.49 & 1.48 & 2.24\tabularnewline
10 & 1.35 & 1.34 & 1.85 & 1.49 & 1.47 & 2.27\tabularnewline
11 & 1.44 & 1.42 & 2.16 & 1.54 & 1.52 & 2.43\tabularnewline
12 & 1.30 & 1.29 & 1.73 & 1.48 & 1.46 & 2.23\tabularnewline
13 & 1.39 & 1.37 & 1.96 & 1.48 & 1.46 & 2.22\tabularnewline
14 & 1.38 & 1.36 & 1.94 & 1.48 & 1.46 & 2.22\tabularnewline
15 & 1.31 & 1.29 & 1.74 & 1.53 & 1.52 & 2.38\tabularnewline
16 & 1.29 & 1.28 & 1.68 & 1.43 & 1.41 & 2.07\tabularnewline
17 & 1.37 & 1.36 & 1.96 & 1.46 & 1.44 & 2.19\tabularnewline
18 & 1.41 & 1.40 & 2.00 & 1.51 & 1.49 & 2.31\tabularnewline
19 & 1.46 & 1.44 & 2.23 & 1.51 & 1.49 & 2.37\tabularnewline
20 & 1.46 & 1.44 & 2.23 & 1.51 & 1.49 & 2.30\tabularnewline
21 & 1.35 & 1.34 & 1.90 & 1.49 & 1.47 & 2.27\tabularnewline
22 & 1.41 & 1.39 & 2.08 & 1.49 & 1.47 & 2.28\tabularnewline
23 & 1.35 & 1.34 & 1.93 & 1.46 & 1.44 & 2.21\tabularnewline
24 & 1.45 & 1.43 & 2.12 & 1.55 & 1.53 & 2.41\tabularnewline
25 & 1.44 & 1.42 & 2.07 & 1.53 & 1.52 & 2.34\tabularnewline
26 & 1.34 & 1.33 & 1.89 & 1.43 & 1.42 & 2.13\tabularnewline
27 & 1.46 & 1.45 & 2.15 & 1.50 & 1.49 & 2.30\tabularnewline
all & 1.31 (min 0.00, max 2.31) & & 1.81 & 1.41 (min $-$0.01, max 2.41) & & 2.08\tabularnewline
\bottomrule
\end{tabular}}\ifdim\wd\tblbox>\linewidth\resizebox{\linewidth}{!}{\usebox{\tblbox}}\else\usebox{\tblbox}\fi
\endgroup
\caption*{\textbf{Table C.11a:} The learned QK-normalisation gains at step 44,000 (mean over the three seeds of the mean gain per 128-dimensional vector, the Dense and Mixed columns being the mean gains \(g_q\), \(g_k\) of each arm; \(g_q g_k\) is the mean elementwise product, a per-dimension weight on the query--key product; descriptive, since the RoPE rotation mixes each dimension pair). The last row pools every value of every layer and seed (minimum and maximum over values).}
\end{table*}

\begin{table*}[tp]
\centering
\begingroup\footnotesize\setlength{\tabcolsep}{4pt}
\sbox{\tblbox}{\begin{tabular}{@{}lllll@{}}
\toprule
Arm & Layers & \(g_q g_k\), highest (0--15) & \(g_q g_k\), high-mid (16--31) & \(g_q g_k\), low-mid (32--47)\tabularnewline
\midrule
Dense & 0 & 0.12 & 0.14 & 0.31\tabularnewline
Dense & 1--27 & 1.37 & 1.78 & 2.12\tabularnewline
Mixed & 0 & 0.06 & 0.09 & 0.33\tabularnewline
Mixed & 1--27 & 1.51 & 2.04 & 2.47\tabularnewline
\bottomrule
\end{tabular}}\ifdim\wd\tblbox>\linewidth\resizebox{\linewidth}{!}{\usebox{\tblbox}}\else\usebox{\tblbox}\fi
\endgroup
\caption*{\textbf{Table C.11b:} (columns 1/2) Frequency selectivity of the learned gains: pair-averaged \(g_q g_k\) by RoPE-frequency quarter (rotate-half pairing, dimension \emph{i} with \emph{i} + 64, index 0 the highest frequency; means over seeds and, for layers 1--27, over layers). The correlation column is between the frequency index (0--63) and the layer-averaged pair product, with the number of layers in which the lowest-frequency quarter exceeds the highest; the CV column is the median over seeds and layers of the coefficient of variation of the product across the 128 dimensions.}
\end{table*}
\begin{table*}[tp]
\centering
\begingroup\footnotesize\setlength{\tabcolsep}{4pt}
\sbox{\tblbox}{\begin{tabular}{@{}lllll@{}}
\toprule
Arm & Layers & \(g_q g_k\), lowest (48--63) & corr(frequency index, product) & median CV over dimensions\tabularnewline
\midrule
Dense & 0 & 0.40 & -- & --\tabularnewline
Dense & 1--27 & 2.19 & 0.91 (26/27 layers) & 0.24\tabularnewline
Mixed & 0 & 0.42 & -- & --\tabularnewline
Mixed & 1--27 & 2.57 & 0.93 (26/27 layers) & 0.28\tabularnewline
\bottomrule
\end{tabular}}\ifdim\wd\tblbox>\linewidth\resizebox{\linewidth}{!}{\usebox{\tblbox}}\else\usebox{\tblbox}\fi
\endgroup
\caption*{\textbf{Table C.11b:} (columns 2/2) Frequency selectivity of the learned gains: pair-averaged \(g_q g_k\) by RoPE-frequency quarter (rotate-half pairing, dimension \emph{i} with \emph{i} + 64, index 0 the highest frequency; means over seeds and, for layers 1--27, over layers). The correlation column is between the frequency index (0--63) and the layer-averaged pair product, with the number of layers in which the lowest-frequency quarter exceeds the highest; the CV column is the median over seeds and layers of the coefficient of variation of the product across the 128 dimensions.}
\end{table*}

\begin{table*}[tp]
\centering
\begingroup\footnotesize\setlength{\tabcolsep}{4pt}
\sbox{\tblbox}{\begin{tabular}{@{}lllll@{}}
\toprule
Model & Mean, all 113 & Mean, Q/K & Mean, residual/final & Changed, all\tabularnewline
\midrule
Dense s41 & 0.9170 & 1.2832 & 0.8571 & 99.1\%\tabularnewline
Mixed s41 & 0.9420 & 1.3943 & 0.8679 & 98.7\%\tabularnewline
Dense s42 & 0.9290 & 1.2987 & 0.8684 & 99.0\%\tabularnewline
Mixed s42 & 0.9605 & 1.3941 & 0.8895 & 99.0\%\tabularnewline
Dense s43 & 0.9167 & 1.2933 & 0.8551 & 99.2\%\tabularnewline
Mixed s43 & 0.9503 & 1.4012 & 0.8764 & 98.5\%\tabularnewline
\bottomrule
\end{tabular}}\ifdim\wd\tblbox>\linewidth\resizebox{\linewidth}{!}{\usebox{\tblbox}}\else\usebox{\tblbox}\fi
\endgroup
\caption*{\textbf{Table C.11c:} Learned gains under the standard recipe (step 44,000; 113 fp32 gain tensors per checkpoint, 56 Q/K and 57 residual/final; ``changed'' = \textbar g $-$ 1\textbar{} \textgreater{} 2$^{-8}$). The learnable-gain ALiBi arms (seeds 42, 43) have 56 Q/K tensors with mean 1.311 and 1.304, 99.5\% and 99.3\% changed, first-layer means 1.015 and 1.012 (\path{FINAL7_20261023_gain_stats.tsv}).}
\end{table*}

\section{Figures}\label{figures}

\begin{figure*}[tp]
\centering
\includegraphics[width=\textwidth]{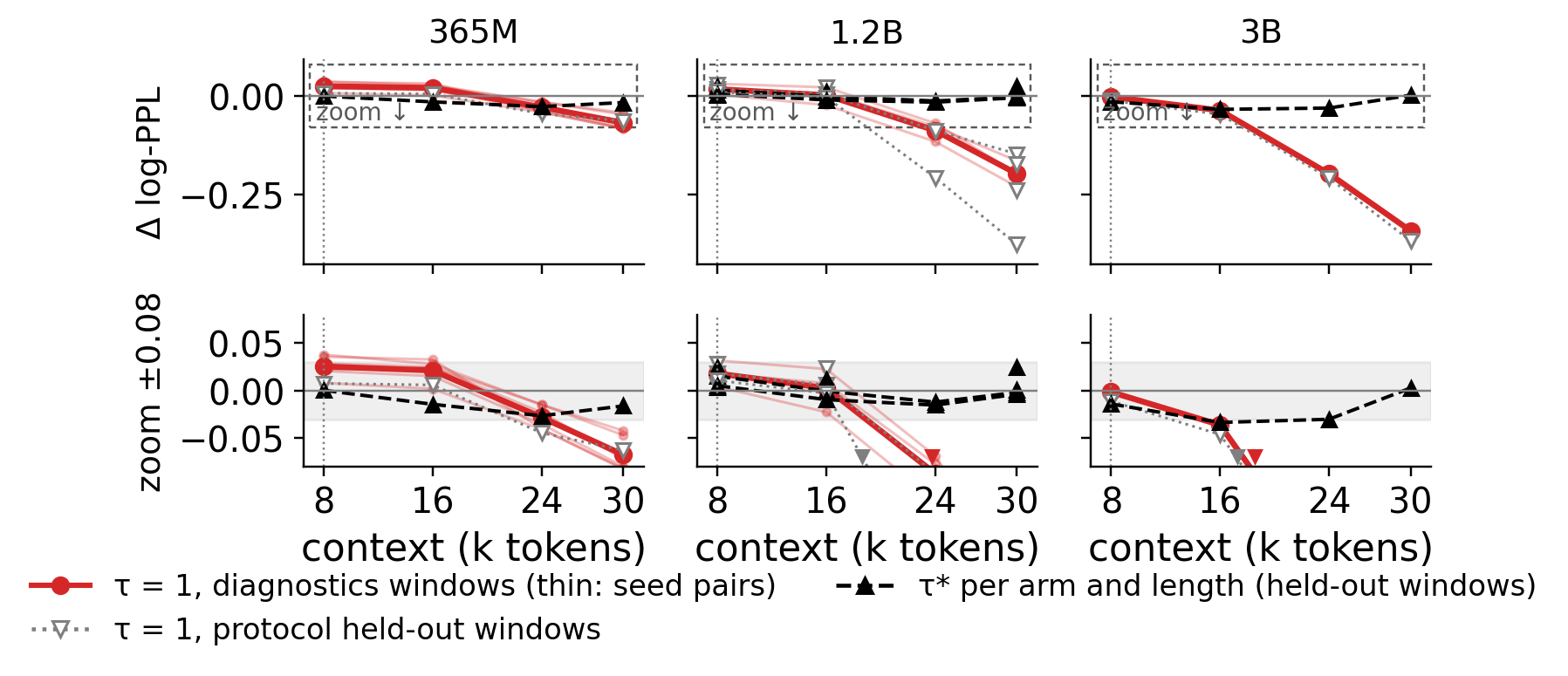}
\caption{\textbf{Context extension under full attention, default versus calibrated temperature} (canonical frozen-gain pairs: 6/3/1 at $\tau$~=~1; 1/4/1 calibrated). Red: $\Delta$~=~log PPL(Mixed) $-$ log PPL(Dense) at $\tau$~=~1 on the diagnostics windows, negative when Mixed is better (thin: seed pairs; thick: mean). Grey hollow triangles: $\tau$~=~1 on the protocol's held-out windows of the calibrated pairs; black dashed: the same windows at each arm's selected $\tau^{*}$. Bottom: the dashed box magnified; band: $\pm$0.03; arrowheads: a series leaves the panel; dotted line: the 8k training length. Lines connect series with all four lengths (1.2B seeds 70--72: markers only).}
\end{figure*}

\section{Reproducibility package}\label{reproducibility-package}

The reproducibility package, released with the result files, contains the analysis scripts (\path{analysis_lib.py}, \path{build_tables.py}, \path{check_tables.py}, \path{make_figures.py}, \path{check_figures.py}, the per-run table scripts \path{resolution_20260921.py}, \path{alibi_20260921.py}, \path{regime_20260922.py}, \path{final_20260924.py}, \path{final3_final4_20261005.py}, \path{final567_20261028.py}), the canonical cell table (\path{cells_20260917.csv.gz}) and every later run's per-window result rows and notes (\path{*_eval_all_rows.tsv}, \path{NOTE_*.md}), the gain statistics that were exported (\path{FINAL7_20261023_gain_stats.tsv}, \path{FINAL3_20260927_gain_stats_from_notes.tsv}, \path{tables/provenance_final567.json}), the frozen-baseline cells re-scored before the standard-recipe run, and \path{tables/} with every generated table and \path{numbers_*.json}. Every number in the main text and every plotted value traces to a table and from there to a cell or a result row; \path{check_short_v3.py} is the script that verified the main text against the tables. The training and evaluation code is at \url{https://github.com/rubentium/SparseAtt}; the archive (about 13 MB compressed) is available from the author.

\end{document}